\documentclass[sigconf]{acmart}

\AtBeginDocument{%
  }

\copyrightyear{2023}
\acmYear{2023}
\setcopyright{acmlicensed}\acmConference[CIKM '23]{Proceedings of the 32nd ACM International Conference on Information and Knowledge Management}{October 21--25, 2023}{Birmingham, United Kingdom}
\acmBooktitle{Proceedings of the 32nd ACM International Conference on Information and Knowledge Management (CIKM '23), October 21--25, 2023, Birmingham, United Kingdom}
\acmPrice{15.00}
\acmDOI{10.1145/3583780.3614997}
\acmISBN{979-8-4007-0124-5/23/10}

\usepackage{hyperref}       % hyperlinks
\usepackage{url}            % simple URL typesetting
\usepackage{xcolor}         % colors
\usepackage{amsmath,mathtools}
\usepackage{graphicx}
\usepackage{algorithm} 
\usepackage{multirow}
\usepackage{caption, makecell}
\usepackage{threeparttable}
\usepackage{enumitem}
\usepackage{listings}
\usepackage{wrapfig}
\usepackage{breqn}
\usepackage{pifont}
\usepackage{tabularx}

\usepackage{tikz}
\usetikzlibrary{positioning}
\usepackage[noend]{algpseudocode}

\newcommand{\etal}{\textit{et al}. }
\newcommand{\ie}{\textit{i}.\textit{e}., }

\DeclareMathOperator{\diag}{\operatorname{diag}}

\DeclareMathOperator*{\argmin}{\operatorname{arg\,min}}

\DeclareMathOperator{\softmax}{\operatorname{softmax}}

\DeclareMathOperator{\vol}{\operatorname{vol}}
\DeclareMathOperator{\sort}{\operatorname{sort}}

\newtheorem{theorem}{Theorem}
\newtheorem{lemma}[theorem]{Lemma}
\newtheorem{corollary}[theorem]{Corollary}
\newtheorem{definition}{Definition}

\begin{document}

\title{On the Trade-off between Over-smoothing and Over-squashing in Deep Graph Neural Networks}


\author{Jhony H. Giraldo}
\affiliation{%
  \institution{LTCI, Télécom Paris - Institut Polytechnique de Paris}
  \streetaddress{19 Place Marguerite Perey}
  \city{Palaiseau}
  \country{France}}
\email{jhony.giraldo@telecom-paris.fr}

\author{Konstantinos Skianis}
\affiliation{%
  \institution{BLUAI}
  %\institution{BLUAI}
  \streetaddress{}
  \city{Athens}
  \country{Greece}}
\email{skianis.konstantinos@gmail.com}

\author{Thierry Bouwmans}
\affiliation{%
  \institution{Laboratoire MIA, La Rochelle Université}
  \streetaddress{}
  \city{La Rochelle}
  \country{France}}
\email{tbouwman@univ-lr.fr}

\author{Fragkiskos D. Malliaros}
\affiliation{%
  \institution{Université Paris-Saclay, CentraleSupélec, Inria}
  \streetaddress{}
  \city{Gif-sur-Yvette}
  \country{France}}
\email{fragkiskos.malliaros@centralesupelec.fr}

\renewcommand{\shortauthors}{Giraldo et al.}

\begin{abstract}  
  Graph Neural Networks (GNNs) have succeeded in various computer science applications, yet deep GNNs underperform their shallow counterparts despite deep learning's success in other domains.
  Over-smoothing and over-squashing are key challenges when stacking graph convolutional layers, hindering deep representation learning and information propagation from distant nodes.
  Our work reveals that over-smoothing and over-squashing are intrinsically related to the spectral gap of the graph Laplacian, resulting in an inevitable trade-off between these two issues, as they cannot be alleviated simultaneously.
  To achieve a suitable compromise, we propose adding and removing edges as a viable approach.
  We introduce the Stochastic Jost and Liu Curvature Rewiring (SJLR) algorithm, which is computationally efficient and preserves fundamental properties compared to previous curvature-based methods.
  Unlike existing approaches, SJLR performs edge addition and removal during GNN training while maintaining the graph unchanged during testing.
  Comprehensive comparisons demonstrate SJLR's competitive performance in addressing over-smoothing and over-squashing.
  
  % Graph Neural Networks (GNNs) have shown success in various computer science applications.
  % However, while deep learning architectures have excelled in other domains, deep GNNs still underperform their shallow counterparts.
  % Over-smoothing and over-squashing are two main challenges when stacking multiple graph convolutional layers, where GNNs struggle to learn deep representations and propagate information from distant nodes.
  % Our work reveals that over-smoothing and over-squashing are intrinsically related to the spectral gap of the graph Laplacian.
  % Consequently, there is a trade-off between these two issues as it is impossible to alleviate both simultaneously.
  % We argue that a viable approach is to add and remove edges in order to achieve a suitable compromise.
  % To this end, we propose the Stochastic Jost and Liu Curvature Rewiring (SJLR) algorithm, which offers a less computationally expensive solution compared to previous curvature-based rewiring methods, while preserving fundamental properties.
  % Unlike existing methods, SJLR performs edge addition and removal during the training phase of GNNs, while keeping the graph unchanged during testing.
  % A comprehensive comparison of SJLR with previous techniques for addressing over-smoothing and over-squashing is performed, where the proposed algorithm shows competitive performance.
\end{abstract}

\begin{CCSXML}
<ccs2012>
   <concept>
       <concept_id>10010147.10010257.10010321</concept_id>
       <concept_desc>Computing methodologies~Machine learning algorithms</concept_desc>
       <concept_significance>500</concept_significance>
       </concept>
   <concept>
       <concept_id>10010520.10010521.10010542.10010294</concept_id>
       <concept_desc>Computer systems organization~Neural networks</concept_desc>
       <concept_significance>500</concept_significance>
       </concept>
 </ccs2012>
\end{CCSXML}

\ccsdesc[500]{Computing methodologies~Machine learning algorithms}
\ccsdesc[500]{Computer systems organization~Neural networks}

\keywords{Graph neural networks, over-smoothing, over-squashing, curvature}


\maketitle

\section{Introduction}

\begin{figure}
    \centering
    \includegraphics[width=0.95\columnwidth]{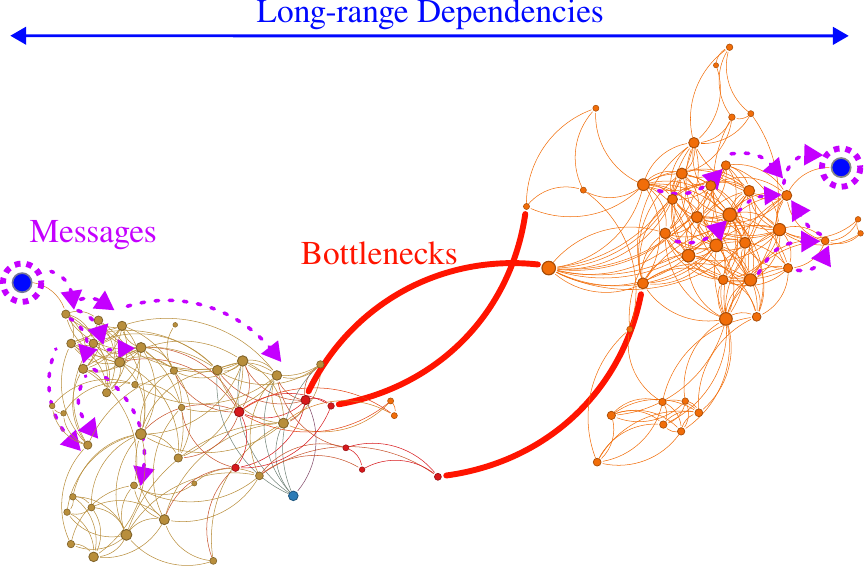}
    \caption{Long-range dependencies in graph neural networks.}
    \label{fig:teaser}
\end{figure}

Graph representation learning is a growing research area, offering a versatile tool for modeling structured data.
In this context, Graph Neural Networks (GNNs) have gained considerable attention from the research community \cite{bruna2014spectral,kipf2017semi,velickovic2018graph,hamilton2017inductive,wu2020comprehensive}.
GNNs extend Convolutional Neural Networks (CNNs) \cite{lecun2015deep} to graph-structured data, enabling powerful models to capture complex dependencies between graph nodes.
%enabling the development of powerful models that can capture complex dependencies between graph nodes.
GNNs find applications in diverse domains, including semi-supervised learning \cite{kipf2017semi}, social network analysis \cite{uwents2011neural}, misinformation detection \cite{benamira2019semisupervised}, materials modeling \cite{pmlr-v202-duval23a}, drug discovery \cite{zitnik2017predicting,gainza2020deciphering}, and computer vision \cite{li2019deepgcns,han2022vision,chen2022survey,giraldo2021graph,prummel2023inductive}.

Although GNNs show great promise in modeling graph-structured data, they are not immune to common neural network limitations, such as over-fitting and vanishing gradients \cite{li2019deepgcns}. 
Two intrinsic limitations of GNNs, over-smoothing \cite{li2018deeper} and over-squashing \cite{alon2021bottleneck}, remain poorly understood.
These issues arise when stacking multiple graph convolutional layers, leading to degraded node representations and distorted information from distant nodes.
Over-smoothing results from node features becoming more similar with increasing convolutional layers \cite{li2018deeper}, while over-squashing occurs due to large information compression through bottleneck edges \cite{alon2021bottleneck}.
These issues are particularly relevant in graphs with large diameters and long-range dependencies between nodes \cite{dwivedi2022long}, as illustrated in Fig. \ref{fig:teaser}, where an exponentially growing number of messages is compressed as they traverse through bottleneck edges.
Various methods have been proposed to address these challenges \cite{oono2020graph,rong2020dropedge,chien2021adaptive,liu2023curvdrop,alon2021bottleneck,topping2022understanding}, yet their relationship remains unformally analyzed \cite{karhadkar2023fosr}.

In this work, we establish a fundamental topological relationship between over-smoothing and over-squashing in deep GNNs.
We leverage the properties of the random walk matrix and the spectral gap of the Laplacian matrix to investigate the phenomenon of over-smoothing, demonstrating how node representations exponentially converge to a stationary distribution \cite{chung1997spectral}.
Similarly, building upon the insights from Topping \etal \cite{topping2022understanding}, we establish a close connection between over-squashing and the spectral gap.
We employ the Cheeger inequality \cite{cheeger1970lower} to highlight the inherent trade-off between over-smoothing and over-squashing, emphasizing that improving one aspect invariably worsens the other.
To navigate this trade-off, we propose the \textit{Stochastic Jost and Liu Curvature Rewiring} (SJLR) algorithm.
The proposed algorithm introduces the Jost and Liu Curvature (JLC), a computationally efficient approximation of Ollivier's Ricci curvature \cite{ollivier2009ricci} as presented in \cite{jost2014ollivier}.
Unlike previous methods, SJLR dynamically adds and removes edges during GNN training to mitigate both over-smoothing and over-squashing, while ensuring the graph remains unchanged during evaluation.
Notably, the JLC metric offers a less computationally complex alternative to the Balanced Forman Curvature (BFC) proposed by Topping \etal \cite{topping2022understanding}, while preserving important theoretical properties.
To evaluate the effectiveness of SJLR, we conduct extensive benchmarking experiments on both homophilous and heterophyllous graph datasets.
The results demonstrate the competitive performance of SJLR in addressing over-smoothing and over-squashing, highlighting its potential as a promising solution in the field of deep GNNs.

This work makes the following main contributions: 
\begin{enumerate}
    \item The establishment of a significant topological relationship between over-smoothing and over-squashing, offering valuable theoretical insights into the behavior of deep GNNs.
    \item The introduction of SJLR, a novel rewiring algorithm specifically designed to address both over-smoothing and over-squashing.
    Notably, SJLR stands out as the first algorithm to perform edge removal and addition exclusively during training, ensuring that the original graph remains unchanged.
    \item Extensive experimentation to evaluate the effectiveness and properties of SJLR.
    Through comprehensive benchmarking and analysis, we provide empirical evidence supporting the performance and benefits of SJLR in mitigating over-smoothing and over-squashing issues.
\end{enumerate}
% \begin{enumerate}
%     \item We establish a topological relationship between over-smoothing and over-squashing, providing relevant theoretical insights into deep GNNs.
%     \item We propose a novel rewiring algorithm, SJLR, to alleviate both over-smoothing and over-squashing.
%     SJLR is the first algorithm where the removal and addition of edges are performed only during training, so the original graph is not modified at all.
%     \item We conduct extensive experiments to evaluate SJLR's effectiveness and properties.
% \end{enumerate}
The remainder of the paper is structured as follows: Section \ref{sec:related_work} reviews related work. Section \ref{sec:background} introduces mathematical notation and preliminary concepts. Section \ref{sec:oversmoothing_oversquashing_tradeoff} presents the relationship between over-smoothing and over-squashing. Section \ref{sec:JLC_algorithm} describes the SJLR algorithm. Finally, we report our experimental results in Section \ref{sec:experiments_results} and present concluding remarks in Section \ref{sec:conclusions}.

%1) we establish a topological relationship between over-smoothing and over-squashing; 2) we propose a novel rewiring algorithm, SJLR, to alleviate both over-smoothing and over-squashing; 3) we conduct extensive experiments to evaluate SJLR's effectiveness and properties; and 4) we release a codebase that facilitates training, hyperparameter tuning, and evaluation of methods for mitigating over-smoothing or over-squashing. Appendix \ref{app:codebase} provides additional details on the codebase. The remainder of the paper is structured as follows: Section \ref{sec:related_work} reviews related work. Section \ref{sec:background} introduces mathematical notation and preliminary concepts. Section \ref{sec:oversmoothing_oversquashing_tradeoff} presents the relationship between over-smoothing and over-squashing. Section \ref{sec:JLC_algorithm} describes the SJLR algorithm. Finally, we report our experimental results in Section \ref{sec:experiments_results} and present concluding remarks in Section \ref{sec:conclusions}.

\section{Related Work}
\label{sec:related_work}

%\subsection{Over-smoothing}
%\textcolor{red}{Update...}

Over-smoothing, which refers to the problem of node embeddings of distinct classes becoming indistinguishable when stacking multiple layers in GNNs, was first discussed in \cite{li2018deeper}.
Since then, several methods have been proposed to alleviate over-smoothing, which can be classified into the following categories:
%Over-smoothing was first discussed in \cite{li2018deeper} to explain why node embeddings of distinct classes become indistinguishable when stacking multiple layers in GNNs.
%Later, several methods were proposed to ease over-smoothing.
%We can briefly classify these approaches as:
\begin{enumerate}[leftmargin=*]
    \item \textbf{Graph rewiring methods:} Klicpera \etal \cite{klicpera2019predict,klicpera2019diffusion} proposed an improved propagation scheme based on PageRank for rewiring the graph as a pre-processing step.
    Similarly, Rong \etal \cite{rong2020dropedge}, Huang \etal \cite{huang2022towards}, and Liu \etal \cite{liu2023curvdrop} introduced methods that drop edges from the graph when training the GNN.
    Finally, Chen \etal \cite{chen2020measuring} presented an approach that adaptively changes the graph topology. 
    %\cite{klicpera2019predict,klicpera2019diffusion} proposed an improved propagation scheme based on PageRank for rewiring the graph as a pre-processing step. \cite{rong2020dropedge} and \cite{huang2022towards} introduced methods that drop edges from the graph when training the GNN. Finally, \cite{chen2020measuring} presented an approach that adaptively changes the graph topology.
    \item \textbf{Normalization techniques:} Zhao \etal \cite{zhao2020pairnorm} and Zhou \etal \cite{zhou2020towards} proposed node-embeddings normalization techniques to address over-smoothing directly, \ie they tried to avoid nodes becoming indistinguishable. Similarly, Oono and Suzuki \cite{oono2020graph} presented a procedure that normalizes the weights of the GNN architecture.
    %\cite{zhao2020pairnorm} and \cite{zhou2020towards} proposed node-embeddings normalization techniques to address over-smoothing directly, \ie they tried to avoid nodes becoming indistinguishable. Similarly, \cite{oono2020graph} presented a procedure that normalizes the weights of the GNN architecture.
    \item \textbf{Architectural changes:} 
    Li \etal \cite{li2019deepgcns} proposed dilated convolutions and residual/dense connections in GNNs to create deep architectures. Chen \etal \cite{chen2020iterative} introduced an iteratively learning graph structure and graph embedding such that their method learns a better graph structure based on better node embeddings, and vice versa. Chien \etal \cite{chien2021adaptive} presented a new graph convolutional filter inspired by the graph signal processing literature \cite{sandryhaila2014discrete} to avoid over-smoothing.
    %\cite{li2019deepgcns} proposed dilated convolutions and residual/dense connections in GNNs to create deep architectures. \cite{chen2020iterative} introduced an iteratively learning graph structure and graph embedding such that their method learns a better graph structure based on better node embeddings, and vice versa. \cite{chien2021adaptive} presented a new graph convolutional filter inspired by the graph signal processing literature \cite{sandryhaila2014discrete} to avoid over-smoothing.
    \item \textbf{Subgraphs approaches:} Zeng \etal \cite{zeng2021decoupling} proposed to train GNNs of arbitrary depths with localized subgraphs.
    %\cite{zeng2021decoupling} proposed to train GNNs of arbitrary depths with localized subgraphs.
\end{enumerate}

%\subsection{Over-squashing}

Recent studies have highlighted over-squashing as a critical issue that hampers the ability of GNNs to effectively propagate information between distant nodes in a graph \cite{alon2021bottleneck}.
%Recent studies have identified over-squashing as a new issue that limits the ability of GNNs to propagate information between distant nodes in the graph \cite{alon2021bottleneck}. 
Alon and Yahav \cite{alon2021bottleneck} proposed a rewiring method in which a fully-adjacent matrix is added in the last GNN layer to mitigate this problem.
%A rewiring method was proposed by Alon and Yahav \cite{alon2021bottleneck}, where a fully-adjacent matrix is added in the last GNN layer to address this problem.
Topping \etal \cite{topping2022understanding} and Di Giovanni \etal \cite{di2023over} further contributed to the understanding of over-squashing, offering theoretical insights into its origins, topological alleviation, and the impact of GNN design choices. 
Specifically, Topping \etal \cite{topping2022understanding} introduced a rewiring method based on concepts from Ricci flow curvature in differential geometry \cite{hamilton1988ricci}. 
However, these studies \cite{alon2021bottleneck,topping2022understanding} solely focused on addressing over-squashing and did not consider the trade-off between over-smoothing and over-squashing.

% Topping \etal \cite{topping2022understanding} and Di Giovanni \etal \cite{di2023over} further proposed theoretical works explaining the origins of over-squashing, how it can be alleviated from a topological perspective, and the impact of some design choices of GNNs on over-squashing.
% Particularly, Topping \etal \cite{topping2022understanding} proposed a rewiring method using concepts from Ricci flow curvature in differential geometry \cite{hamilton1988ricci}.
% However, the studies \cite{alon2021bottleneck} and \cite{topping2022understanding} focused solely on addressing over-squashing and did not consider the trade-off between over-smoothing and over-squashing.

More recently, Karhadkar \etal \cite{karhadkar2023fosr} and Liu \etal \cite{liu2023curvdrop} proposed empirical methods to alleviate both over-smoothing and over-squashing.
Karhadkar \etal \cite{karhadkar2023fosr} acknowledged the trade-off between over-squashing and over-smoothing but did not provide formal proof of its existence.
Furthermore, their method solely involved edge addition.
Liu \etal \cite{liu2023curvdrop} also presented a method to address both problems, but they only focused on edge removal based on a curvature metric.
Differently to these works, 1) we formally prove the existence of the trade-off, 2) we argue that both edge addition and removal are necessary, and 3) we aim to explore the relationship between these two issues, providing a more comprehensive understanding of their interplay.

% More recently, Karhadkar \etal \cite{karhadkar2023fosr} and Liu \etal \cite{liu2023curvdrop} proposed two empirical methods to alleviate over-smoothing and over-squashing.
% Karhadkar \etal \cite{karhadkar2023fosr} mentioned the trade-off between over-squashing and over-smoothing, but they did not prove its existence.
% Similarly, their method only add edges.
% Liu \etal \cite{liu2023curvdrop} also proposed a method to alleviate both problems, but they only remove edges based a curvature metric.
% In this work, 1) we prove the existence of the trade-off, 2) we argue that adding and removing edges is required, and 3) we aim to explore the relationship between these two issues while providing a more comprehensive understanding of their interplay.

Transformers offer an alternative approach to address over-smoothing and over-squashing and have gained increasing interest in the graph and computer vision domains, with several studies exploring their effectiveness \cite{yun2019graph,cai2020graph,chen2022survey}.
Notably, Ying \etal \cite{ying2021transformers} observed that transformer architectures are less susceptible to over-smoothing compared to GNNs.
Additionally, Kreuzer \etal \cite{kreuzer2021rethinking} explained that transformers avoid over-squashing due to the presence of direct paths between distant nodes.
However, transformers have significant computational and memory limitations as each node attends to all other vertices, making them less suitable for large-scale graph applications.
Moreover, improper training of transformers can result in a mixture of local and non-local interactions.
In this work, we propose a novel approach to simultaneously address both over-smoothing and over-squashing.
We utilize JLC and node-embedding metrics to rewire the graph, without relying on attention mechanisms, offering potential advantages for large-scale graph problems.

\section{Notation and Background}
\label{sec:background}

In this paper, calligraphic letters like $\mathcal{V}$ designate sets, and $\vert \mathcal{V} \vert$ represents their cardinality.
Uppercase boldface letters such as $\mathbf{A}$ represent matrices, and lowercase boldface letters like $\mathbf{x}$ denote vectors.
$\diag(\mathbf{x})$ is a diagonal matrix with entries $x_1,x_2,\dots,x_n$.
$\Vert \cdot \Vert$ is the $\ell_2$-norm of a vector and $(\cdot)^{\mathsf{T}}$ represents transposition.
$\mathbf{I}$ is the identity matrix, and $\mathbf{1}$ is a vector of ones with appropriate dimensions.
Finally, $\mathcal{N}_{i}$ is the set of neighbors of node $i$.

\subsection{Preliminaries} 

A graph is a mathematical entity represented as $G=(\mathcal{V},\mathcal{E})$, where $\mathcal{V}=\{1,\dots,N\}$ is the set of $N$ nodes and the set of edges is ${\mathcal{E}\subseteq \{(i,j)\mid i,j\in \mathcal{V}\;{\textrm {and}}\;i\neq j\}}$ such that $(i,j)$ is an edge between the vertices $i$ and $j$. %is the set of edges such that $(i,j)$ is an edge between the vertices $i$ and $j$.
In this paper, we consider undirected, connected, and unweighted graphs.
$\mathbf{A} \in \{0,1\}^{N\times N}$ is the adjacency matrix of $G$ such that $\mathbf{A}(i,j)= 1$ if $(i,j)\in \mathcal{E}$ and $\mathbf{A}(i,j)= 0$ otherwise.
%As a consequence, $\mathbf{A}$ is symmetric for undirected graphs.
Moreover, $\mathbf{D} \in \mathbb{R}^{N\times N}$ is the diagonal degree matrix of $G$ such that $\mathbf{D}(i,i)=\sum_{j=1}^N \mathbf{A}(i,j)~\forall~i = 1,\dots,N$, and $d_i=\mathbf{D}(i,i)$.
$\mathbf{L} = \mathbf{D}-\mathbf{A}$ is the positive semi-definite combinatorial Laplacian operator.
Similarly, $\mathbf{L}_{\text{sym}}=\mathbf{D}^{-\frac{1}{2}}\mathbf{L}\mathbf{D}^{-\frac{1}{2}}=\mathbf{I}-\mathbf{D}^{-\frac{1}{2}}\mathbf{A}\mathbf{D}^{-\frac{1}{2}}$ is the symmetrically normalized Laplacian matrix with eigenvalues $0=\lambda_1 < \lambda_2 \leq \dots \leq \lambda_N \leq 2$ and corresponding eigenvectors $\{ \mathbf{u}_1,\mathbf{u}_2,\dots,\mathbf{u}_N\}$.

\subsection{Cheeger Inequality and Cheeger Constant} 

\begin{definition}
    \label{def:Cheeger_constant}
    Let $\mathcal{S} \subset \mathcal{V}$ be a subset of nodes of $G$.
    Let $\partial \mathcal{S}$ be the set of edges going from a node in $\mathcal{S}$ to a node in $\mathcal{V} \setminus \mathcal{S}$, \ie $\partial \mathcal{S} \triangleq \left\{ \{u,v\} \in \mathcal{E} : u \in \mathcal{S}, v \in \mathcal{V} \setminus \mathcal{S} \right\}$.
    Therefore, we can define the Cheeger constant $h_G$ of $G$ as $h_G \triangleq \min_{\mathcal{S}} h_G(\mathcal{S})$, where $h_G(\mathcal{S})=\vert \partial \mathcal{S} \vert / \min (\vol(\mathcal{S}),\vol(\mathcal{V} \setminus \mathcal{S}))$, and $\vol(\mathcal{S})=\sum_{i \in \mathcal{S}} d_i$.
    %\begin{equation}
    %    h_G \triangleq \min_{\mathcal{S}} h_G(\mathcal{S}),
    %    \label{eqn:cheeger_constant}
    %\end{equation}
    %where $h_G(\mathcal{S})=\vert \partial \mathcal{S} \vert / \min (\vol(\mathcal{S}),\vol(\mathcal{V} \setminus \mathcal{S}))$, $\vol(\mathcal{S})=\sum_{i \in \mathcal{S}} d_i$.
    %\begin{equation}
    %    \partial \mathcal{S} \triangleq \left\{ \{u,v\} \in \mathcal{E} : v \in \mathcal{S}, v \in \mathcal{V} \setminus \mathcal{S} \right\}.
    %    \label{eqn:edge_boundary}
    %\end{equation}
\end{definition}
Intuitively, the Cheeger constant in Definition \ref{def:Cheeger_constant} is small when there exists a \textit{bottleneck} in $G$, \ie when there are two sets of nodes with few edges between them.
Similarly, we know that $h_G > 0$ iff $G$ is a connected graph \cite{chung1997spectral}.
We can relate the Cheeger constant $h_G$ with the first non-zero eigenvalue of $\mathbf{L}_{\text{sym}}$ through the Cheeger inequality:
\begin{equation}
    2h_G \geq \lambda_2 \geq \frac{h_G^2}{2}.
    \label{eqn:cheeger_inequality}
\end{equation}
We notice from \eqref{eqn:cheeger_inequality} that for having less ``bottleneckness'' in the graph, we need to promote big values of $h_G$, \ie having large values of $\lambda_2$ will increase $h_G$ since $h_G \geq \lambda_2/2$.

%we need to promote big values of $\lambda_2$ since $h_G \geq \lambda_2/2$.

%However, promoting big values of $\lambda_2$ accelerates the convergence to the stationary distribution in the random walk adjacency matrix as shown in Lemma \ref{lem:distance_to_stationary_distribution}.

\subsection{Message Passing Neural Networks (MPNNs)} 

Let $G$ be a graph with a set of input features $\mathbf{X} \in \mathbb{R}^{N \times F_1} = [\mathbf{x}_1,\dots, \mathbf{x}_N]^{\mathsf{T}}$.
The output of a generic MPNN is defined as follows \cite{gilmer2017neural}:
\begin{equation}
    \mathbf{h}_i^{(l+1)} = \phi_l \left( \mathbf{h}_i^{(l)}, \sum_{j=1}^{N} \hat{\mathbf{A}}(i,j) \psi_l (\mathbf{h}_i^{(l)},\mathbf{h}_j^{(l)}) \right),
    \label{eqn:message_passing_NN}
\end{equation}
where $\mathbf{H}^{(l)} \in \mathbb{R}^{N \times F_l} = [\mathbf{h}^{(l)}_1,\dots, \mathbf{h}^{(l)}_N]^{\mathsf{T}}$ is the $F_l$-dimensional embeddings after $l$ layers such that each $\mathbf{h}^{(l)}_i \in \mathbb{R}^{F_l}$ and $\mathbf{H}^{(1)}=\mathbf{X}$, $\psi_l:\mathbb{R}^{F_l} \times \mathbb{R}^{F_l} \to \mathbb{R}^{F'_l}$ is a family of message functions, $\hat{\mathbf{A}}$ is an augmented normalized adjacency matrix, and $\phi_l : \mathbb{R}^{F_l} \times \mathbb{R}^{F'_l} \to \mathbb{R}^{F_{l+1}}$ is an update function.

\subsection{Over-smoothing}

Graph convolutions usually use smoothing functions on each layer.
Therefore, when we apply several graph convolution layers, the performance can suffer from over-smoothing, where node embeddings from different clusters become mixed up.
Over-smoothing lacks a formal definition in the literature \cite{zhao2020pairnorm}.
However, we can think of over-smoothing as a random walk transition matrix that is repeatedly applied to a node feature, thus converging to a stationary distribution and washing away all the feature information (this convergence is explained in \eqref{eqn:convergence_to_stationay_distribution}, Section \ref{sec:stationary_distribution}).

\subsection{Over-squashing}

Over-squashing is a more recent and less understood problem than over-smoothing.
A graph learning problem has \textit{long-range dependencies} when the outputs of GNNs depend on features of interacting distant nodes.
In that scenario, information from non-adjacent nodes should be propagated through the network without distortion.
Let $\mathcal{B}_r \triangleq \{j \in \mathcal{V}: d_G(i,j) \leq r\}$ be the receptive field of an $r$-layer GNN, where $d_G$ is the shortest-path distance and $r \in \mathbb{N}$.
Let $\partial \mathbf{h}^{(r)}_i/\partial \mathbf{x}_j$ be the \textit{Jacobian} of a node embedding $\mathbf{h}^{(r)}_i$ with respect to some input feature $\mathbf{x}_j$ in node $j$.
Over-squashing can be understood as the inability of $\mathbf{h}^{(r)}_i$ to be affected by $\mathbf{x}_j$ at a distance $r$.
Topping \etal \cite{topping2022understanding} proved that $\left| \partial \mathbf{h}^{(r+1)}_i/\partial \mathbf{x}_j \right| \leq (\alpha \beta)^{r+1}\hat{\mathbf{A}}^{r+1}(i,j)$, if $\vert \nabla \phi_l \vert \leq \alpha$ and $\vert \nabla \psi_l \vert \leq \beta$ for $0 \leq l \leq r$, with $\phi_l$, $\psi_l$ differentiable functions.
In many graphs, $\vert \mathcal{B}_r \vert$ grows exponentially with $r$, and then representations of an exponential amount of neighboring nodes should be compressed into fixed-size vectors.
For example, if $d_G(i,j) = r+1$ in a binary tree, we have that $\hat{\mathbf{A}}^{r+1}(i,j) = 2^{-1}3^{-r}$, which gives an exponential decay of the node dependence on input features at distance $r$ \cite{topping2022understanding}.
This phenomenon is referred to as over-squashing of information \cite{alon2021bottleneck,topping2022understanding, di2023over}.

\section{Understanding the Over-smoothing vs. Over-squashing Trade-off}
\label{sec:oversmoothing_oversquashing_tradeoff}

\subsection{The Stationary Distribution on Graphs}
\label{sec:stationary_distribution}

Let $\mathbf{P}=\mathbf{D}^{-1}\mathbf{A}$ be the random walk transition matrix.
For any initial distribution $f:\mathcal{V} \to \mathbb{R}$ with $\sum_{v \in \mathcal{V}}f(v)=1$, the distribution after $k$ steps is given by $\mathbf{f}^{\mathsf{T}}\mathbf{P}^k$, where $\mathbf{f}\in \mathbb{R}^{N\times 1}$ is the vector of initial distributions such that $\mathbf{f}(i)$ is the function evaluated on the $i$th node.
The random walk is \textit{ergodic} when there is a unique \textit{stationary distribution} $\boldsymbol{\pi}$ satisfying that $\lim_{s\to \infty} \mathbf{f}^{\mathsf{T}}\mathbf{P}^s = \boldsymbol{\pi}$ \cite{chung1997spectral}.

\begin{lemma}[Chung \cite{chung1997spectral}]
    \label{lem:distance_to_stationary_distribution}
    Let $\mathbf{P}$ be an ergodic random walk transition matrix, where G is connected and non-bipartite, let $\boldsymbol{\pi}$ be its stationary distribution, and let $\mathbf{f}$ be any initial distribution.
    For $s \in \mathbb{N}^+$, we have:
    \begin{equation}
        \label{eqn:convergence_to_stationay_distribution}
        \Vert \mathbf{f}^{\mathsf{T}} \mathbf{P}^s - \boldsymbol{\pi} \Vert \leq e^{-s \lambda'} \frac{\max_i \sqrt{d_i}}{\min_j \sqrt{d_j}},
    \end{equation}
    where $\lambda'=\lambda_2$ if $1-\lambda_2 \geq \lambda_N - 1$, and $2 - \lambda_N$ otherwise.
    Therefore, we can compute the value of $s$ such that $\Vert \mathbf{f}^{\mathsf{T}} \mathbf{P}^s - \boldsymbol{\pi} \Vert \leq \epsilon$ as follows:
    \begin{equation}
        s \geq \frac{1}{\lambda' \log \left( \max_i \sqrt{d_i} / \epsilon \min_j \sqrt{d_j} \right)}.
        \label{eqn:s_steps_to_epsilon}
    \end{equation}
    %where:
    %\begin{equation}
    %    \lambda' = \begin{cases}
    %        \lambda_2 & \text{if } 1-\lambda_2 \geq \lambda_N - 1, \\
    %        2 - \lambda_N & \text{otherwise}.
    %    \end{cases}
    %\end{equation}
    %Therefore, after $s \geq \frac{1}{\lambda'} \log \left( \frac{\max_x \sqrt{\mathbf{D}(x,x)}}{\epsilon \min_y \sqrt{\mathbf{D}(y,y)}} \right)$ steps, $\Vert \mathbf{f}^{\mathsf{T}} \mathbf{P}^s - \boldsymbol{\pi} \Vert_2 \leq \epsilon$.\\
    %Proof: See \cite{chung1997spectral}.
\end{lemma}

Notice that $\lambda'$ is either $\lambda_2$ or $2-\lambda_N$ in Lemma \ref{lem:distance_to_stationary_distribution}.
However, we can show that only $\lambda_2$ is crucial.
Suppose that $\lambda_N - 1 > 1-\lambda_2$, so that $\lambda'=2-\lambda_N$.
We can consider the lazy walk on the graph $G'$ formed by adding a loop of weight $d_i$ to each node $i$, \ie $\mathbf{A}+\mathbf{D}$.
Therefore, the new graph Laplacian has eigenvalues $\tilde{\lambda}_i=\lambda_i/2 \leq 1$ \cite{chung1997spectral}, and then $1-\tilde{\lambda}_2 \geq 1 - \tilde{\lambda}_N \geq 0$.

The key message of Lemma \ref{lem:distance_to_stationary_distribution} is a simplified version of the same observations of \cite{rong2020dropedge,oono2020graph}, \ie GNNs converge exponentially to a stationary distribution when stacking several layers.
However, we show that the convergence of this exponential function depends on the spectral gap $\lambda_2$.
%first non-zero eigenvalue of the graph $\lambda_2$ (the spectral gap).
We use this result later to show the underlying relationship between over-smoothing and over-squashing.
Similarly, we can have a simplified explanation of why sparsification methods in GNNs, like DropEdge \cite{rong2020dropedge}, can alleviate over-smoothing.
\begin{lemma}[Chung \cite{chung1997spectral}]
    \label{lem:upper_bound_lambda_2}
    Let $G$ be a graph with diameter $Di \geq 4$, then:%and let $\rho$ denote the maximum degree of $G$, then:
    \begin{equation}
        \lambda_2 \leq 1-2 \frac{\sqrt{(\max_i d_i) -1}}{\max_i d_i}\left(1-\frac{2}{Di}\right)+\frac{2}{Di}.
    \end{equation}
    %Proof: See \cite{nilli1991second}.
\end{lemma}
Lemma \ref{lem:upper_bound_lambda_2} shows an upper bound of $\lambda_2$ so that reducing the maximum degree of $G$ promotes small values of $\lambda_2$, \ie having a sparser graph promotes low values of $\lambda_2$ and thus high values of $s$ in \eqref{eqn:s_steps_to_epsilon}.

\subsection{Over-smoothing and Over-squashing}
\label{sec:oversmoothing_oversquashing_in_GNNs}

We can establish a link between the Cheeger constant $h_G$ and the parameter $s$ in \eqref{eqn:s_steps_to_epsilon} as follows:
\begin{theorem}
    \label{trm:cheeger_constant_and_mixing_time}
    Let $h_G$ be the Cheeger constant of $G$, and let $s$ be the number of required steps such that the $\ell_2$ distance between $\mathbf{\mathbf{f}^{\mathsf{T}} \mathbf{P}^s}$ and $\boldsymbol{\pi}$ is at most $\epsilon$.
    Therefore, we have that:
    \begin{equation}
        2h_G \geq \frac{1}{s} \log \left( \frac{\max_i \sqrt{d_i}}{\epsilon \min_j \sqrt{d_j}} \right).
    \end{equation}
    Proof: see Appendix \ref{app:proof_cheeger_constant_and_mixing_time}.
\end{theorem}

From Theorem \ref{trm:cheeger_constant_and_mixing_time}, we have that if $s \to 0$ then $h_G \to \infty$, \ie we can promote less “bottleneckness” in the graph if we accelerate the convergence to the stationary distribution. Similarly, if $h_G \to 0$ then $s \to \infty$, \ie we can avoid converging to the stationary distribution if we promote a bottleneck-kind structure in the graph.
In other words, we can reduce over-squashing by accelerating the convergence to the stationary distribution that worsens over-smoothing.
Correspondingly, we can avoid over-smoothing by promoting having bottlenecks that worsen over-squashing.
We also prove in Appendix \ref{app:mixing_time} that 1) if $s \to \infty$ then $h_G \to 0$, and 2) if $h_G \to \infty$ then $s \to 0$, so $s \to \infty \iff h_G \to 0$ and $h_G \to \infty \iff s \to 0$.

We can make the connection between over-smoothing and over-squashing more precisely using Theorem \ref{trm:cheeger_constant_and_mixing_time}, the Simple Graph Convolution (SGC) model \cite{wu2019simplifying}, and the developments in \cite{topping2022understanding}.
In other words, we can use an SGC with a random walk kernel to show how the node embeddings converge to the stationary distribution when we stack several layers according to Theorem \ref{trm:cheeger_constant_and_mixing_time}.
Similarly, Topping \etal \cite{topping2022understanding} explained how reducing bottlenecks in the graphs can alleviate over-squashing, \ie the receptive field of each node in a deep GNN will be polynomial in the hop-distance rather than exponential.
SGC is a simplified version of the Graph Convolutional Network (GCN) \cite{kipf2017semi}, where we remove all projection parameters and all non-linear activation functions between layers.
We use SGC as a proxy of GNNs, as in \cite{zhao2020pairnorm}, to study the relationship between over-smoothing and over-squashing.
As a consequence, our theoretical results are only available for linear-based GNNs like SGC \cite{wu2019simplifying} or simple spectral graph convolution \cite{zhu2021simple} (experimental results on other GNNs are presented in Section \ref{sec:experiments_results}).
We leave for future work the analysis of the relationship between over-smoothing and over-squashing for more complex GNNs architectures.%with non-linear activation functions.

\begin{figure}
    \centering
    \includegraphics[width=\columnwidth]{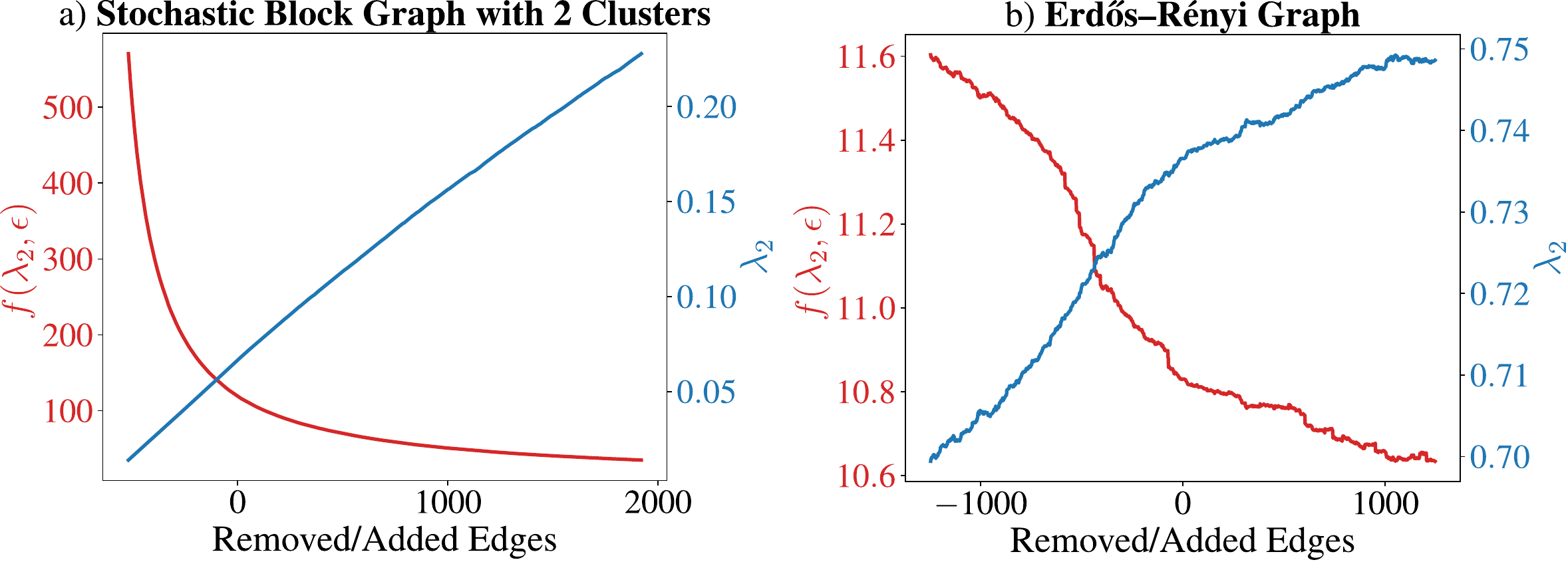}
    \caption{Mixing steps $f(\lambda_2,\epsilon)$ for $\epsilon=5 \times 10^{-4}$ vs. number of removed or added edges for a) one stochastic block model graph with two clusters, and b) one Erdős-Rényi graph.}
    \label{fig:oversmoothing_vs_oversquashing}
\end{figure}

Let $f(\lambda_2,\epsilon)=\frac{1}{\lambda_2} \log \left( \max_i \sqrt{d_i}/\epsilon \min_j \sqrt{d_j} \right)$ be the mixing steps of our graph, \ie the lower bound in the maximum number of layers of an SGC such that the difference between the initial and stationary distribution is at most $\epsilon$.
Figure \ref{fig:oversmoothing_vs_oversquashing} shows, for $\epsilon=5 \times 10^{-4}$, how $f(\lambda_2,\epsilon)$ and $\lambda_2$ change when we add or remove edges in one artificial stochastic block model graph and one Erdős-Rényi graph.
We can increase the mixing steps by removing edges, \ie we can alleviate over-smoothing by making the graph more ``bottleneckness''.
This partially explains why DropEdge \cite{rong2020dropedge} can alleviate over-smoothing.
On the other hand, we increase $\lambda_2$ by adding edges as shown in Fig. \ref{fig:oversmoothing_vs_oversquashing}, so we promote higher values of $h_G$, \ie we can alleviate over-squashing by making the graph less ``bottleneckness''.
This can partially explain why the methodology by Topping \etal \cite{topping2022understanding} can alleviate over-squashing.
However, there is a trade-off between $f(\lambda_2,\epsilon)$ and $\lambda_2$ from a topological point of view, \ie we can increase $f(\lambda_2,\epsilon)$ by removing key edges but $\lambda_2$ will decrease, and vice versa.
The algorithm to add and remove edges is explained in Section \ref{sec:JLC_algorithm}.

\section{Curvature Rewiring Algorithm}
\label{sec:JLC_algorithm}

\begin{figure*}
    \centering
    \includegraphics[width=\textwidth]{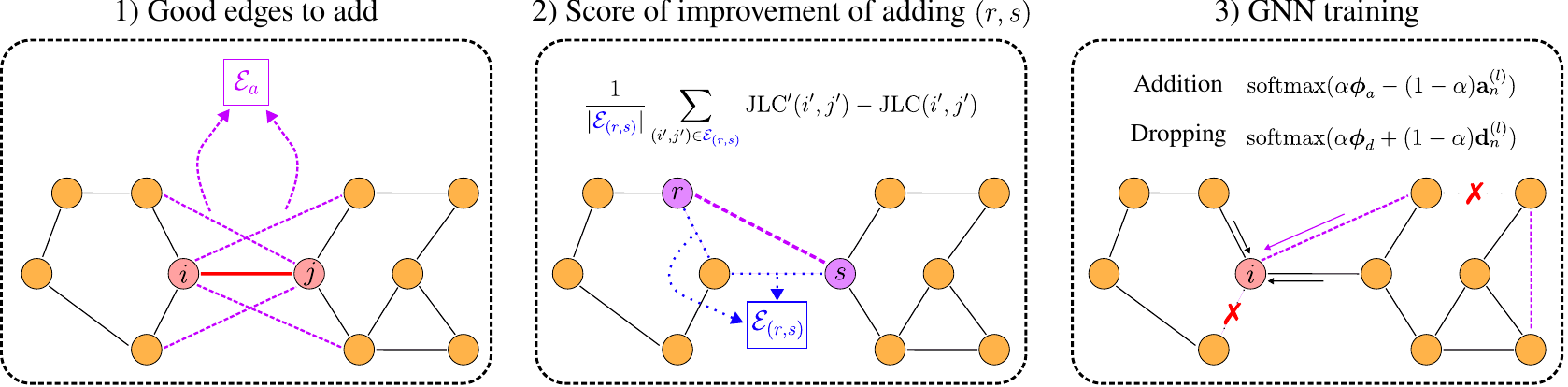}
    \caption{The pipeline of the proposed Stochastic Jost and Liu curvature Rewiring (SJLR) algorithm. SJLR first saves potential good edges $\mathcal{E}_a$ to add to the graph, based on the JLC metric. Secondly, SJLR computes the score of improvement of adding these new edges as the average of the JLC improvement that concerns these edges. Thirdly, edges are added and removed according to some probability distribution during the graph neural network training. The graph remains unchanged during testing.}
    \label{fig:pipeline}
\end{figure*}

%as a pre-processing step, and later edges are removed while training the GNN using JLC and node-embeddings information.
%We keep the addition and removal of edges in two different stages attempting to find a good solution in the over-smoothing vs. over-squashing trade-off.

Topping \etal \cite{topping2022understanding} proposed a method to alleviate over-squashing using concepts of Ricci flow curvature.
We can understand curvature-based methods using the following analysis:
\begin{theorem}[Lin \etal \cite{lin2011ricci}]
    \label{trm:positive_ricci_curvature}
    Let $G$ be a finite graph, let $\lambda_2$ be its spectral gap, and let $\kappa(i,j)$ be the Ricci curvature as defined in \cite{lin2011ricci}.
    If for any edge $(i,j)$, $\kappa(i,j) \geq \kappa > 0$, then $\lambda_2 \geq \kappa$.
    %Proof: See \cite{lin2011ricci}.
\end{theorem}
\begin{corollary}
    \label{cor:relationship_ricci_cheeger_constant}
    If $\kappa(i,j) \geq \kappa > 0$ for any edge $(i,j)$, then $2h_G \geq \kappa$.\\
    Proof: Using the Cheeger inequality and Theorem \ref{trm:positive_ricci_curvature}, we have that $2h_G \geq \lambda_2 \geq \kappa$, and then $2h_G \geq \kappa$.
\end{corollary}
From Theorem \ref{trm:positive_ricci_curvature} and Corollary \ref{cor:relationship_ricci_cheeger_constant}, we can conclude that if we have positive Ricci curvature everywhere, then $2h_G \geq \kappa$.
Therefore, increasing curvature will make the graph less ``bottleneckness''.
In this paper, we use a bound of the Ollivier's Ricci curvature \cite{ollivier2009ricci} presented in \cite{jost2014ollivier}.
\begin{definition}[Jost and Liu Curvature (JLC) \cite{jost2014ollivier}]
    For any edge $(i,j)$ in a finite graph:    
    \begin{dmath}
        \text{\normalfont JLC}(i,j) = -\left( 1 - \frac{1}{d_i} - \frac{1}{d_j} - \frac{\#(i,j)}{d_i \wedge d_j} \right)_+ - \left( 1 - \frac{1}{d_i} - \frac{1}{d_j} - \frac{\#(i,j)}{d_i \vee d_j} \right)_+ + \frac{\#(i,j)}{d_i \vee d_j},
    \end{dmath}
    where $\#(i,j)$ is the number of triangles which include $(i,j)$ as nodes, $c_+ \triangleq \max(c,0)$, $c \vee t \triangleq \max(c,t)$, and $c \wedge t \triangleq \min(c,t)$.
\end{definition}
\begin{theorem}[Jost and Liu \cite{jost2014ollivier}]
    \label{trm:JLC}
    On a locally finite graph we have that $\kappa(i,j) \geq \text{\normalfont JLC}(i,j)$.
    %Proof: See \cite{jost2014ollivier}.
\end{theorem}
\begin{corollary}
    \label{cor:JLC_positive}
    If $\, \text{\normalfont JLC}(i,j) \geq \kappa > 0$ for any edge $(i,j)$, then $\kappa(i,j) \geq \kappa > 0$.\\
    Proof: Using Theorems \ref{trm:positive_ricci_curvature} and \ref{trm:JLC} we have that $\kappa(i,j) \geq \text{\normalfont JLC}(i,j) \geq \kappa \geq 0$, then $\kappa(i,j) \geq \kappa \geq 0$.
\end{corollary}
From Corollary \ref{cor:JLC_positive}, we can conclude that if we have positive JLC everywhere in $G$, $\kappa(i,j)$ will also be positive everywhere.
As a result, having positive JLC ensures that the receptive field of each node in a deep GNN will be polynomial in the hop-distance rather than exponential (see Corollary 3 in \cite{topping2022understanding}).
Topping \etal \cite{topping2022understanding} defined the Balanced Forman Curvature (BFC) metric to solve over-squashing.
However, BFC requires counting triangles, 4-cycles, and the maximal number of 4-cycles traversing a common node for each edge.
This process makes BFC very computationally intensive and unsuitable for practical applications.
Thus, JLC is less computationally complex than BFC while keeping the same theoretical properties about the polynomial receptive field growth.

\subsection{Stochastic Jost and Liu Curvature Rewiring Details}

Our algorithm uses JLC and node feature information to perform rewiring.
Topping \etal \cite{topping2022understanding} proposed a method where edges are added and removed as a pre-processing step.
However, their reason for removing edges was not properly justified.
We argue that deleting edges is also important to alleviate over-smoothing, according to the developments in Section \ref{sec:oversmoothing_oversquashing_tradeoff}.
SJLR also adds and removes edges but is fundamentally different.
We stochastically add and remove edges only during training, so we can alleviate over-squashing and over-smoothing without modifying the initial graph, maintaining its original properties.
%perform addition and removal of edges
%We keep both processes separate so that we can alleviate over-squashing and over-smoothing according to the dataset, as shown in Fig. \ref{fig:pipeline}.
We define a set of hyperparameters for SJLR: 1) let $p_A$ be the percentage of added edges, 2) let $p_D$ be the percentage of dropped edges, and 3) let $\alpha \in [0,1]$ be a variable controlling how important is the JLC metric against the embedding information while dropping or adding edges.
%3) let $\tau$ be a variable controlling how stochastic SJLR is when adding edges, and 4) let $\alpha \in [0,1]$ be a variable controlling how important is the JLC metric against the embedding information while dropping edges.
We optimize the hyperparameters in the validation set so that SJLR can choose either if the specific dataset requires more addition or removal of edges, \ie SJLR tries to find the ``sweet point'' in the trade-off between over-smoothing and over-squashing\footnote{For further details about the ``sweet point'' see Appendix \ref{app:sween_point}.}.

\algrenewcommand\algorithmicrequire{\textbf{Input:}}
\algrenewcommand\algorithmicensure{\textbf{Initialization:}}
\begin{algorithm}[t]
\begin{algorithmic}[1]
\Require graph $G$, GNN architecture 
\Ensure hyperparameters $p_A$, $p_D$, $\alpha$, $\rho=1$
        \State $\mathcal{E}_a$, JLC = \Call{Bank\_Edges}{$G$}
        \State $\boldsymbol{\phi}_d$, $\boldsymbol{\phi}_a$ = \Call{Scores\_Improvement}{$G$, $\mathcal{E}_a$, JLC}
        \For{each layer $l$ in GNN while training}
            \State Compute $\mathbf{d}^{(l)}(p) = \Vert \mathbf{h}^{(l)}_i - \mathbf{h}^{(l)}_j \Vert~\forall~(i,j) \in \mathcal{E}$, $1 \leq p \leq \vert \mathcal{E} \vert$ \label{lst:Eucl_dist_drop}
            \State Drop $p_D \vert \mathcal{E} \vert$ edges from $\mathcal{E}$ according to the probability \newline \hspace*{2em} distribution $\softmax(\alpha \boldsymbol{\phi}_d + (1-\alpha) \mathbf{d}_n^{(l)})$ \label{lst:drop}
            \State Compute $\mathbf{a}^{(l)}(q) = \Vert \mathbf{h}^{(l)}_r - \mathbf{h}^{(l)}_s \Vert~\forall~(r,s) \in \mathcal{E}_a$, $1 \leq q \leq \vert \mathcal{E}_a \vert$ \label{lst:Eucl_dist_adding}
            \State Add $p_A \vert \mathcal{E} \vert$ edges from $\mathcal{E}_a$ according to the probability \newline \hspace*{2em} distribution $\softmax(\alpha \boldsymbol{\phi}_a - (1-\alpha) \mathbf{a}_n^{(l)})$ \label{lst:adding}
        \EndFor \vspace{0.2cm}
        \Function{Bank\_Edges}{Graph $G$}
            \State Compute $\text{JLC}(i,j)~\forall~(i,j) \in \mathcal{E}$
            \State $\text{JLC}_{\text{sorted}} = \sort(\text{JLC})$, sort the JLC scores in ascending order
            \State Create an empty set of potential good edges to add $\mathcal{E}_a = \emptyset$
            \While{$\vert \mathcal{E}_a \vert < 2p_A \vert \mathcal{E} \vert$}
                \State $(i', j')$ is the $\rho$th edge in $\text{JLC}_{\text{sorted}}$
                \State $\mathcal{A} = \{ \{ (\mathcal{N}_{i'} \setminus j') \times j' \} \bigcup \{ (\mathcal{N}_{j'} \setminus i') \times i' \} : \mathcal{A} \notin \mathcal{E} \}$
                \State $\mathcal{E}_a = \mathcal{E}_a \bigcup \mathcal{A}$, $\rho = \rho + 1$
            \EndWhile
            \State \Return $\mathcal{E}_a$, $\text{JLC}(i,j)~\forall~(i,j) \in \mathcal{E}$
        \EndFunction \vspace{0.2cm}
        \Function{Scores\_Improvement}{Graph $G$, $\mathcal{E}_a$, JLC values}
            \State Create a vector $\boldsymbol{\sigma} \in \mathbb{R}^{\vert \mathcal{E}_a \vert}$
            \For{$m=1$ until $m=\vert \mathcal{E}_a \vert$}
                \State Compute the set of edges $\mathcal{E}_{(r,s)} \subset \mathcal{E}$ that form a triangle \newline \hspace*{4em} with $(r,s)$, where $(r,s)$ is the $m$th edge in $\mathcal{E}_a$
                \State Create a new graph $G' = (\mathcal{V}, \{ \mathcal{E} \bigcup (r,s) \} )$
                \State Compute $\text{JLC}'(i',j')~\forall~(i',j') \in \mathcal{E}_{(r,s)}$ in graph $G'$
                \State $\boldsymbol{\sigma}(m) = \frac{1}{\vert \mathcal{E}_{(r,s)} \vert} \sum_{(i',j') \in \mathcal{E}_{(r,s)}} \text{JLC}'(i',j') - \text{JLC}(i',j')$
            \EndFor
            \State Normalize $\text{JLC}(i,j)$ and $\boldsymbol{\sigma}(m)$ to be in $[0,1]$, $\forall$ $(i,j) \in \mathcal{E}$ and \newline \hspace*{2em} $1 \leq m \leq \vert \mathcal{E}_a \vert$ \label{lst:normalization}
            \State \Return $\boldsymbol{\phi}_d$, $\boldsymbol{\phi}_a$
            %\State \Return \textit{sliceAtual}
        \EndFunction
        % \If {$i\geq maxval$}
        %     \State $i\gets 0$
        % \Else
        %     \If {$i+k\leq maxval$}
        %         \State $i\gets i+k$
        %     \EndIf
        % \EndIf
        \end{algorithmic}
    \caption{Stochastic Jost and Liu Curvature Rewiring (SJLR)}
    \label{alg:SJLR}
\end{algorithm}

\begin{table*}[]
\centering
\caption{Statistics of the datasets in the experimental framework of this work.}
\label{tbl:statistics_datasets}
\begin{tabular}{r|ccccccccc}
\toprule
         & \textbf{Cornell} & \textbf{Texas} & \textbf{Wisconsin} & \textbf{Chameleon} & \textbf{Squirrel} & \textbf{Actor} & \textbf{Cora} & \textbf{Citeseer} & \textbf{Pubmed} \\
\midrule
$H(G)$ & $0.11$ & $0.06$ & $0.16$ & $0.25$ & $0.22$ & $0.24$ & $0.83$ & $0.72$ & $0.79$ \\
Nodes & $140$ & $135$ & $184$ & $832$ & $2,186$ & $4,388$ & $2,485$ & $2,120$ & $19,717$ \\
Edges & $219$ & $251$ & $362$ & $12,355$ & $65,224$ & $21,907$ & $5,069$ & $3,679$ & $44,324$ \\
Features & $1,703$ & $1,703$ & $1,703$ & $2,325$ & $2,089$ & $932$ & $1,433$ & $3,703$ & $500$ \\
Classes & $5$ & $5$ & $5$ & $5$ & $5$ & $5$ & $7$ & $6$ & $3$ \\
Directed & \checkmark & \checkmark & \checkmark & \checkmark & \checkmark & \checkmark &  &  &  \\
\bottomrule
\end{tabular}
\end{table*}

Figure \ref{fig:pipeline} shows the pipeline of the proposed Stochastic Jost and Liu curvature Rewiring (SJLR) algorithm, where edges are added and removed while training the GNN using JLC and node-embeddings information.
Algorithm \ref{alg:SJLR} presents our SJLR approach in detail.
The algorithm has as input an initial graph $G$ and a given GNN architecture.
SJLR first computes a bank of potential good edges to add $\mathcal{E}_a$ for improving the JLC curvature (first part in Fig. \ref{fig:pipeline}), where the JLC metric is calculated (for all $(i,j) \in \mathcal{E}$) and sorted (in ascending order).
%computes and sorts (in ascending order) the JLC for all $(i,j) \in \mathcal{E}$.
%Therefore, a set $\mathcal{E}_a$ is created with potential good edges to add for improving the JLC curvature (first part in Fig. \ref{fig:pipeline}).
To this end, we look at every edge $(i',j')$ from the sorted JLC vector, and we compute $\mathcal{A} = \{ \{ (\mathcal{N}_{i'} \setminus j') \times j' \} \bigcup \{ (\mathcal{N}_{j'} \setminus i') \times i' \} : \mathcal{A} \notin \mathcal{E} \}$, which is the set of edges that form triangles with $(i',j')$ and are not in $\mathcal{E}$.
We append this set $\mathcal{A}$ to $\mathcal{E}_a$ until we have at least $2p_A \vert \mathcal{E} \vert$ edges in $\mathcal{E}_a$.
Therefore, we associate a score $\boldsymbol{\sigma}(m)$ to every edge $(r,s) \in \mathcal{E}_a$, which is computed as the average improvement of curvature from adding that edge $(r,s)$ to the graph (second part in Fig. \ref{fig:pipeline}).
Let $\boldsymbol{\sigma} \in \mathbb{R}^{\vert \mathcal{E}_a \vert}$ be the vector of JLC improvements, such that:
\begin{equation}
    \boldsymbol{\sigma}(m) = \frac{1}{\vert \mathcal{E}_{(r,s)} \vert} \sum_{(i',j') \in \mathcal{E}_{(r,s)}} \text{JLC}'(i',j') - \text{JLC}(i',j'),
\end{equation}
where $\text{JLC}'(i',j')$ is the JLC metric of edge $(i',j')$ computed in the augmented graph $G' = (\mathcal{V}, \{ \mathcal{E} \cup (r,s) \})$, and $\mathcal{E}_{(r,s)} \subset \mathcal{E}$ is the set of edges that form a triangle with the edge $(r,s) \in \mathcal{E}_a$.
%and $\boldsymbol{\sigma} \in \mathbb{R}^{\vert \mathcal{E}_a \vert}$ is the vector of JLC improvements if adding an edge from $\mathcal{A}$.
%Each $\boldsymbol{\sigma}(m)$ is computed as the average improvement from 
%Similarly, $\text{JLC}'(i',j')$ is the JLC metric of edge $(i',j')$ computed in the augmented graph $G' = (\mathcal{V}, \{ \mathcal{E} \cup (r,s) \})$.
%where $m$ is the iterator index while adding edges.
Before the GNN training loop, we normalize $\text{JLC}(i,j)$ and $\boldsymbol{\sigma}(m)$ as shown in line \ref{lst:normalization}, and save them in two vectors $\boldsymbol{\phi}_d$ and $\boldsymbol{\phi}_a$.
During the GNN training for each layer $l$, 1) we compute the euclidean distances $\mathbf{d}^{(l)}$ and $\mathbf{a}^{(l)}$ as shown in lines \ref{lst:Eucl_dist_drop} and \ref{lst:Eucl_dist_adding}, 2) we normalize $\mathbf{d}^{(l)}$ and $\mathbf{a}^{(l)}$ to be in $[0,1]$ and get $\mathbf{d}_n^{(l)}$ and $\mathbf{a}_n^{(l)}$, and 3) we drop and add edges according to the probabilities distribution in lines \ref{lst:drop} and \ref{lst:adding} (third part in Fig. \ref{fig:pipeline}).
%$\boldsymbol{\phi} \in \mathbb{R}^{\vert \mathcal{E}_A \vert}$ is the vector with the normalized values of $\text{JLC}(i,j)~\forall~(i,j) \in \mathcal{E}_A$, and $\mathbf{d}_n^{(l)} \in [0,1]^{\vert \mathcal{E}_A \vert}$ is the vector of normalized distances between node embeddings.
%The SJLR algorithm is divided into two parts: 1) the addition of edges using the JLC metric, and 2) the removal of edges while training the GNN.
It is worth noting that the addition of potential good edges could be partially parallelized since there is not a sequential procedure.
This is an important difference regarding previous methods \cite{topping2022understanding}, where edges cannot be added in parallel.
SJLR is agnostic to the GNN architecture.
%Notice that SJLR is general to any GNN architecture.
However, we only test SJLR using SGCs \cite{wu2019simplifying} and GCNs \cite{kipf2017semi} in Section \ref{sec:experiments_results}. %due to limited computational resources.
%We also propose an additional instance of our algorithm called SJLR+.
%This rewiring algorithm adds edges as a pre-processing step and removes edges while training like in SJLR.
%However, SJLR+ has two additional architectural changes: 1) we add residual/dense connections in the GNN, and 2) we add PairNorm on each layer.

\section{Experimental Framework and Results}
\label{sec:experiments_results}

%This work strives to understand the underlying relationship between over-smoothing and over-squashing rather than beating the performance of state-of-the-art methods.

\begin{table*}[]
\caption{Comparison results of the proposed SJLR algorithm with several state-of-the-art methods to alleviate over-smoothing and over-squashing with the SGC model as backbone.}
\label{tbl:results_experiment_SGC}
\resizebox{\textwidth}{!}{
\begin{threeparttable}
\begin{tabular}{r|ccccccccc|c}
\toprule
\textbf{Method} & \textbf{Cornell} & \textbf{Texas} & \textbf{Wisconsin} & \textbf{Chameleon} & \textbf{Squirrel} & \textbf{Actor} & \textbf{Cora} & \textbf{Citeseer} & \textbf{Pubmed} & \textbf{\textit{Overall}} \\
\midrule
Baseline & $53.40_{\pm 2.11}$ & $56.69_{\pm 1.78}$ & $47.90_{\pm 1.73}$ & $38.40_{\pm 0.69}$ & $40.52_{\pm 0.54}$ & $29.93_{\pm 0.16}$ & $76.94_{\pm 1.31}$ & $67.45_{\pm 0.80}$ & $71.79_{\pm 2.13}$ & $53.67$ \\
GDC \cite{klicpera2019diffusion} & $58.65_{\pm 1.43}$ & $57.42_{\pm 0.74}$ & $45.93_{\pm 1.05}$ & $38.13_{\pm 0.55}$ & $36.63_{\pm 0.31}$ & \color{red} $\textbf{32.25}_{\pm 0.17}$ & $76.02_{\pm 1.70}$ & $66.22_{\pm 1.13}$ & $71.91_{\pm 2.30}$ & $53.68$ \\
DE \cite{rong2020dropedge} & \color{blue} $\underline{\textit{61.99}}_{\pm 1.04}$ & \color{blue} $\underline{\textit{57.88}}_{\pm 0.81}$ & \color{blue} $\underline{\textit{54.78}}_{\pm 0.89}$ & $40.38_{\pm 0.47}$ & $41.28_{\pm 0.32}$ & $30.62_{\pm 0.17}$ & $80.59_{\pm 0.80}$ & \color{red} $\textbf{68.63}_{\pm 0.51}$ & $74.47_{\pm 1.65}$ & \color{blue} $\underline{
\textit{56.74}}$ \\
PN \cite{zhao2020pairnorm} & $53.11_{\pm 1.36}$ & $50.47_{\pm 1.04}$ & $48.72_{\pm 1.65}$ & \color{red} $\textbf{41.49}_{\pm 0.68}$ & $39.72_{\pm 0.33}$ & $22.58_{\pm 0.29}$ & $75.55_{\pm 0.42}$ & $64.16_{\pm 0.41}$ & $73.81_{\pm 0.52}$ & $52.18$ \\
DGN \cite{zhou2020towards} & $55.68_{\pm 1.32}$ & $57.42_{\pm 2.59}$ & $50.67_{\pm 2.08}$ & \color{blue} $\underline{\textit{40.99}}_{\pm 0.62}$ & \color{blue} $\underline{\textit{41.72}}_{\pm 0.29}$ & $29.53_{\pm 0.18}$ & \color{blue} $\underline{\textit{80.65}}_{\pm 0.48}$ & $67.65_{\pm 0.59}$ & 
\color{blue} $\underline{\textit{74.95}}_{\pm 0.59}$ & $55.47$ \\
SDRF \cite{topping2022understanding} & $54.68_{\pm 1.29}$ & $55.36_{\pm 1.48}$ & $47.81_{\pm 1.51}$ & $38.07_{\pm 0.77}$ & $39.94_{\pm 0.53}$ & $30.04_{\pm 0.17}$ & $76.04_{\pm 1.69}$ & $67.60_{\pm 0.80}$ & $69.62_{\pm 2.35}$ & $53.24$ \\
FoSR \cite{karhadkar2023fosr} & $53.73_{\pm 1.75}$ & $56.33_{\pm 1.37}$ & $47.82_{\pm 2.14}$ & $38.01_{\pm 0.73}$ & $40.68_{\pm 0.42}$ & $30.11_{\pm 0.18}$ & $78.24_{\pm 0.98}$ & $67.04_{\pm 0.83}$ & $72.76_{\pm 2.35}$ & $53.86$ \\
\midrule
SJLR (ours) & \color{red} $\textbf{67.37}_{\pm 1.64}$ & \color{red} $\textbf{58.40}_{\pm 1.48}$ & \color{red} $\textbf{55.42}_{\pm 0.92}$ & $40.17_{\pm 0.49}$ & \color{red} $\textbf{41.91}_{\pm 0.34}$ & \color{blue} $\underline{\textit{30.81}}_{\pm 0.18}$ & \color{red} $\textbf{81.24}_{\pm 0.77}$ & \color{blue} $\underline{\textit{68.39}}_{\pm 0.69}$ & \color{red} $\textbf{76.28}_{\pm 0.96}$ & \color{red} $\textbf{57.78}$ \\
\bottomrule
\end{tabular}
\begin{tablenotes}
\item The best and second-best performing methods on each dataset are shown in {\color{red}\textbf{red}} and {\color{blue}\textit{\underline{blue}}}, respectively.
%\item \scriptsize \hspace{0.15cm} \textbf{OOM} (Out Of Memory), this experiment was executed on three NVIDIA GeForce RTX 2080.
%\item \scriptsize $^\dagger$ We were unable to reproduce the results of GDC \cite{klicpera2019diffusion} under our experimental framework, so we copied the results directly from \cite{topping2022understanding}.
\end{tablenotes}
\end{threeparttable}
}
\end{table*}

\begin{table*}[]
\caption{Comparison results of the proposed SJLR algorithm with several state-of-the-art methods to alleviate over-smoothing and over-squashing for the GCN model as backbone.}
\label{tbl:results_experiment_GCN}
\resizebox{\textwidth}{!}{
\begin{tabular}{r|ccccccccc|c}
\toprule
\textbf{Method} & \textbf{Cornell} & \textbf{Texas} & \textbf{Wisconsin} & \textbf{Chameleon} & \textbf{Squirrel} & \textbf{Actor} & \textbf{Cora} & \textbf{Citeseer} & \textbf{Pubmed} & \textbf{\textit{Overall}} \\
\midrule
Baseline & \color{blue} $\underline{\textit{67.34}}_{\pm 1.50}$ & $58.05_{\pm 0.96}$ & ${{52.10}}_{\pm 0.95}$ & $40.35_{\pm 0.48}$ & \color{red} $\textbf{42.12}_{\pm 0.29}$ & $28.62_{\pm 0.36}$ & $81.81_{\pm 0.26}$ & $68.35_{\pm 0.35}$ & $78.25_{\pm 0.37}$ & \color{blue} $\underline{\textit{57.44}}$ \\
RDC \cite{li2019deepgcns} & $63.78_{\pm 1.68}$ & $59.47_{\pm 1.00}$ & $50.89_{\pm 1.00}$ & $40.33_{\pm 0.51}$ & \color{blue} $\underline{\textit{41.98}}_{\pm 0.31}$ & $28.97_{\pm 0.33}$ & $81.54_{\pm 0.26}$ & $68.70_{\pm 0.35}$ & $78.42_{\pm 0.39}$ & $57.12$ \\
GDC \cite{klicpera2019diffusion} & $64.18_{\pm 1.36}$ & $56.43_{\pm 1.15}$ & $49.61_{\pm 0.95}$ & $38.49_{\pm 0.51}$ & $33.20_{\pm 0.29}$ & \color{red} $\textbf{31.08}_{\pm 0.27}$ & \color{red} $\textbf{82.63}_{\pm 0.23}$ & $69.15_{\pm 0.30}$ & \color{red} $\textbf{79.04}_{\pm 0.37}$ & $55.98$ \\
DE \cite{rong2020dropedge} & $63.39_{\pm 1.29}$ & $57.41_{\pm 0.93}$ & $47.84_{\pm 0.86}$ & \color{blue} $\underline{\textit{40.80}}_{\pm 0.55}$ & $41.68_{\pm 0.39}$ & \color{blue} $\underline{\textit{29.99}}_{\pm 0.21}$ & $81.90_{\pm 0.24}$ & $68.99_{\pm 0.36}$ & $78.53_{\pm 0.26}$ & $56.73$ \\
PN \cite{zhao2020pairnorm} & $64.44_{\pm 1.39}$ & \color{red} $\textbf{60.93}_{\pm 1.15}$ & $51.78_{\pm 0.95}$ & $40.37_{\pm 0.59}$ & $40.92_{\pm 0.31}$ & $28.21_{\pm 0.21}$ & $78.89_{\pm 0.32}$ & $66.95_{\pm 0.40}$ & $76.60_{\pm 0.41}$ & $56.57$ \\
DGN \cite{zhou2020towards} & $65.19_{\pm 1.79}$ & $58.91_{\pm 0.93}$ & $50.76_{\pm 0.92}$ & $40.06_{\pm 0.60}$ & $41.30_{\pm 0.32}$ & $28.32_{\pm 0.36}$ & $81.34_{\pm 0.31}$ & $69.25_{\pm 0.35}$ & $78.06_{\pm 0.42}$ & $57.02$ \\
FA \cite{alon2021bottleneck} & $53.57_{\pm 0.00}$ & $59.26_{\pm 0.00}$ & $43.02_{\pm 0.49}$ & $27.76_{\pm 0.29}$ & $31.51_{\pm 0.00}$ & $26.69_{\pm 0.50}$ & $29.85_{\pm 0.00}$ & $23.23_{\pm 0.00}$ & $39.24_{\pm 0.00}$ & $37.13$ \\
SDRF \cite{topping2022understanding} & $63.88_{\pm 1.68}$ & $56.40_{\pm 0.89}$ & $40.99_{\pm 0.62}$ & $40.74_{\pm 0.45}$ & $41.44_{\pm 0.37}$ & $28.95_{\pm 0.33}$ & $81.42_{\pm 0.26}$ & \color{blue} $\underline{\textit{69.37}}_{\pm 0.31}$ & $77.74_{\pm 0.42}$ & $55.66$ \\
FoSR \cite{karhadkar2023fosr} & $56.65_{\pm 0.93}$ & $50.01_{\pm 1.37}$ & \color{blue} $\textit{\underline{53.73}}_{\pm 1.08}$ & $40.26_{\pm 0.50}$ & $41.83_{\pm 0.28}$ & $28.80_{\pm 0.35}$ & $81.79_{\pm 0.26}$ & $67.99_{\pm 0.37}$ & $78.26_{\pm 0.39}$ & $55.48$ \\
\midrule
SJLR (ours) & \color{red} $\textbf{71.75}_{\pm 1.50}$ & \color{blue} $\textit{\underline{60.13}}_{\pm 0.89}$ & \color{red} $\textbf{55.16}_{\pm 0.95}$ & \color{red} $\textbf{41.19}_{\pm 0.46}$ & $41.86_{\pm 0.29}$ & $29.89_{\pm 0.20}$ & \color{blue} $\textit{\underline{81.95}}_{\pm 0.25}$ & \color{red} $\textbf{69.50}_{\pm 0.33}$ & \color{blue} $\textit{\underline{78.60}}_{\pm 0.33}$ & \color{red} $\textbf{58.89}$ \\
\bottomrule
\end{tabular}
}
\end{table*}

\subsection{Experiments}
\label{sec:experiments}

We perform a set of experiments to compare SJLR with several approaches in the literature.
%striving to understand the underlying relationship between over-smoothing and over-squashing.
%SJLR is compared to seven state-of-the-art methods to alleviate over-smoothing or over-squashing, including Residual/Dense Connections (RDC) \cite{li2019deepgcns}, Graph Diffusion Convolution (GDC) with personalized PageRank kernel \cite{klicpera2019diffusion}, DropEdge (DE) \cite{rong2020dropedge}, PairNorm (PN) \cite{zhao2020pairnorm}, Differentiable Group Normalization (DGN) \cite{zhou2020towards}, Fully-Adjacent (FA) layers \cite{alon2021bottleneck}, and Stochastic Discrete Ricci Flow (SDRF) \cite{topping2022understanding}.
SJLR is compared to eight state-of-the-art methods to alleviate over-smoothing or over-squashing, including Residual/Dense Connections (RDC) \cite{li2019deepgcns}, Graph Diffusion Convolution (GDC) with personalized PageRank kernel \cite{klicpera2019diffusion}, DropEdge (DE) \cite{rong2020dropedge}, PairNorm (PN) \cite{zhao2020pairnorm}, Differentiable Group Normalization (DGN) \cite{zhou2020towards}, Fully-Adjacent (FA) layers \cite{alon2021bottleneck}, Stochastic Discrete Ricci Flow (SDRF) \cite{topping2022understanding}, and First-order Spectral Rewiring FoSR \cite{karhadkar2023fosr}.
Our GNN base models are SGC \cite{wu2019simplifying} and GCN \cite{kipf2017semi} for all experiments.
We do not test RDC and FA in SGC because this GNN model only has one graph convolutional layer.
We evaluate all methods in nine datasets: Cornell, Texas, and Wisconsin from the WebKB project\footnote{\url{http://www.cs.cmu.edu/afs/cs.cmu.edu/project/theo-11/www/wwkb/}}, Chameleon \cite{rozemberczki2021multi}, Squirrel \cite{rozemberczki2021multi}, Actor \cite{tang2009social}, Cora \cite{mccallum2000automating}, Citeseer \cite{sen2008collective}, and Pubmed \cite{namata2012query}.
We consider the largest connected component of the
graph for each dataset as in \citep{klicpera2019diffusion,topping2022understanding}.
Table \ref{tbl:statistics_datasets} shows the statistics of the datasets tested in this work, where $H(G)$ is the homophily of the graph as defined in \cite{pei2020geom}.
We split the data into a development set and a test set, ensuring that the test set is not used during the hyperparameter optimization process.
%We split the data into train/validation/test sets, where we first divide the data into a development set and a test set.
%This is done once to avoid using the test set in the hyperparameter optimization procedure.
We follow the same experimental framework as in \citep{klicpera2019diffusion,topping2022understanding}, \ie we optimize the hyperparameters for all dataset-preprocessing combinations separately by random search over $100$ data splits.
Furthermore, we report average accuracies on the test set accompanied by $95\%$ confidence intervals calculated by bootstrapping with $1,000$ samples.
For Cora, Citeseer, and Pubmed, the development set contains $1,500$ nodes and the rest of the nodes are used for testing.
Similarly, the train set contains $20$ nodes of each class while the rest of the nodes are used for validation.
As for the other datasets, we use a 60/20/20 split of the nodes, meaning that 60\% of the nodes are assigned for training, 20\% for validation, and 20\% for testing.
%\ie 60\% for training, 20\% for validation, and 20\% for testing.
%We use the same method and random seeds as in \citep{klicpera2019diffusion,topping2022understanding}, so we expect to have the same partitions and comparable results.

% We optimize the hyperparameters with a random search procedure by maximizing the average accuracy in the validation sets.
% The search space and the best hyperparameters for each experiment are provided in Appendix \ref{app:hyperparameters} and \ref{app:experiment_SGC}.
% More details about the experimental framework are provided in Appendix \ref{app:detailed_experiments}.

% The search space for the number of layers for GCN in the first experiment is $\{1,2,3\}$ as in \cite{topping2022understanding} (for further details please see Appendices \ref{app:hyperparameters} and \ref{app:detailed_experiments}), while for SGC we have $\{1,2,3,4\}$.
% However, these search spaces do not provide insights into the problems we could face when stacking multiple layers.
% As a result, we perform a second experiment for the GCN model where we stack several layers in the set $\mathcal{F}=\{2,8,32\}$.
% We follow the same methodology as in the first experiment for each method and each number of layers in $\mathcal{F}$ regarding data partitions and hyperparameters optimization.

\subsection{Implementation Details}

All methods are implemented using PyTorch and PyG \cite{fey2019fast}.
We use the same architectural components in all techniques for a fair comparison.
We use SGC \cite{wu2019simplifying} or GCN \cite{kipf2017semi} as graph convolutional layers.
%notice however that other GNNs like GAT \cite{velickovic2018graph} can be used as well.
We implemented SDRF \cite{topping2022understanding} at our best understanding because there was not an available implementation of the method at the time of conducting the experiments.
However, we use JLC instead of BFC in our implementation of SDRF \cite{topping2022understanding} because of the significant computational resources required to run the hyperparameter optimization using BFC.
%The hyperparameter search space for each method is defined as follows: 1) learning rate $lr \in [0.005, 0.02]$; 2) weight decay $wd \in [0.0001, 0.001]$; 3) hidden units of each graph convolutional layer $hu \in \{16, 32, 64, 128\}$; 4) dropout $d \in [0.3, 0.7]$; 5) the number of layers $L \in \{2,3,4\}$; 6) percentage of added and dropped edges $p_A, p_D \in [0, 1]$; 7) $\alpha \in [0, 1]$; 8) scale $s \in \{0.1, 1, 10, 50, 100\}$ for PN; 9) number of clusters $c \in \{3,4,\dots,10\}$ and balancing factor $bf \in [0.0005, 0.05]$ for DGN; 10) $\alpha_{\text{GDC}} \in [0.01, 0.2]$ and $k \in \{16, 32, 64, 128\}$ for GDC; and 10) stochasticity level $\tau \in [1, 500]$, iterations $it \in \{20, 21, \dots, 4000 \}$, and Ricci curvature upper-bound $C^+ \in [0.1, 40]$ for SDRF.
The hyperparameter search space for each method is defined as follows: 1) learning rate $lr \in [0.005, 0.02]$; 2) weight decay $wd \in [0.0001, 0.001]$; 3) hidden units of each graph convolutional layer $hu \in \{16, 32, 64, 128\}$; 4) dropout $d \in [0.3, 0.7]$; 5) the number of layers $L \in \{2,3,4\}$; 6) percentage of added and dropped edges $p_A, p_D \in [0, 1]$; 7) $\alpha \in [0, 1]$; 8) scale $s \in \{0.1, 1, 10, 50, 100\}$ for PN; 9) number of clusters $c \in \{3,4,\dots,10\}$ and balancing factor $bf \in [0.0005, 0.05]$ for DGN; 10) $\alpha_{\text{GDC}} \in [0.01, 0.2]$ and $k \in \{16, 32, 64, 128\}$ for GDC; 10) stochasticity level $\tau \in [1, 500]$, iterations $it \in \{20, 21, \dots, 4000 \}$, and Ricci curvature upper-bound $C^+ \in [0.1, 40]$ for SDRF; and 11) number of SoFR iterations $itF \in \{1, 2, \dots, 150 \}$.
We use Rectified Linear Unit (ReLU) and log-softmax as activation functions in our GNN architectures.
For GDC, we apply weight decay regularization only in the first graph convolutional layer, otherwise we do not get comparable results as in \cite{klicpera2019diffusion}.
All methods are trained for $1,000$ epochs using Adam optimizer \citep{kingma2015adam}.
We do not use early stopping or learning schedulers for any method.
We make all graphs undirected, and we also remove all the self-loops from the input graph.
%Chebyshev filter size $\text{K} \in \{1,2,3\}$, 7) heads of attention mechanism $hd \in \{1,2,\dots,6\}$, 8) powers for SIGN $pw \in \{1,2,3\}$, 9) $\alpha \in \{1,2,\dots,6\}$, 10) $\epsilon \in [0.5, 2]$, 11) fusion layer $fl \in \{\text{Linear}, \text{MLP}\}$.
%Further details about the implementation details are given in Appendix \ref{app:implementation_details}.
The code is publicly available\footnote{\url{https://github.com/jhonygiraldo/SJLR}} under the MIT license.

\begin{table*}[]
\centering
\caption{Ablation study about dropping and adding edges in SJLR with the SGC model as backbone.}
\label{tbl:ablation}
\resizebox{\textwidth}{!}{
\begin{threeparttable}
\begin{tabular}{cc|ccccccccc}
\toprule
\textbf{Dropping} & \textbf{Adding} & \textbf{Cornell} & \textbf{Texas} & \textbf{Wisconsin} & \textbf{Chameleon} & \textbf{Squirrel} & \textbf{Actor} & \textbf{Cora} & \textbf{Citeseer} & \textbf{Pubmed} \\
\midrule
\ding{55} & \ding{55} & $53.40_{\pm 2.11}$ & $56.69_{\pm 1.78}$ & $47.90_{\pm 1.73}$ & $38.40_{\pm 0.69}$ & $40.52_{\pm 0.54}$ & $29.93_{\pm 0.16}$ & $76.94_{\pm 1.31}$ & $67.45_{\pm 0.80}$ & $71.79_{\pm 2.13}$ \\
\ding{51} & \ding{55} & $59.26_{\pm 1.61}$ & $56.16_{\pm 0.85}$ & $53.64_{\pm 0.97}$ & $39.95_{\pm 0.81}$ & $41.27_{\pm 0.39}$ & $30.04_{\pm 0.19}$ & $80.86_{\pm 0.76}$ & $\textbf{68.48}_{\pm 0.51}$ & $75.19_{\pm 1.51}$ \\
\ding{55} & \ding{51} & $54.07_{\pm 2.36}$ & $56.79_{\pm 1.93}$ & $45.44_{\pm 1.95}$ & $38.20_{\pm 0.58}$ & $41.57_{\pm 0.32}$ & $29.86_{\pm 0.19}$ & $73.90_{\pm 2.01}$ & $66.20_{\pm 1.05}$ & $69.56_{\pm 2.36}$ \\
\ding{51} & \ding{51} & $\textbf{67.37}_{\pm 1.64}$ & $\textbf{58.40}_{\pm 1.48}$ & $\textbf{55.42}_{\pm 0.92}$ & $\textbf{40.17}_{\pm 0.49}$ & $\textbf{41.91}_{\pm 0.34}$ & $\textbf{30.81}_{\pm 0.18}$ & $\textbf{81.24}_{\pm 0.77}$ & $68.39_{\pm 0.69}$ & $\textbf{76.28}_{\pm 0.96}$ \\
\bottomrule
\end{tabular}
\begin{tablenotes}
\item The best result on each dataset are shown in \textbf{bold}.
\end{tablenotes}
\end{threeparttable}
}
\end{table*}

\subsection{Results}

Tables \ref{tbl:results_experiment_SGC} and \ref{tbl:results_experiment_GCN} show the results for SGC and GCN, respectively.
%Table \ref{tbl:results_experiment_non_deep} shows the results of the first experiment related to the random search with few convolutional layers for the GCN model, while the results for SGC are in Appendix \ref{app:experiment_SGC}.
SJLR shows the overall best performance in both cases.
We notice two general trends: 1) rewiring methods such as DE and SJLR dominate in almost all datasets for the experiment with SGC, and 2) GDC leads in the homophilous datasets Cora and Pubmed with GCN.
%heterophyllous datasets like Cornell, Texas, Wisconsin, Chameleon, Squirrel, and Actor; and 2) GDC leads in homophilous datasets like Cora, Citeseer, and Pubmed, while SJLR and FA offer competitive performances.
Our theoretical results are based on the assumption that there are no non-linear activation functions, so perhaps some nuances are missed for GNNs like GCN.
%Similarly, we notice that SJLR outperforms SDRF \cite{topping2022understanding} in all datasets.
Similarly, we notice that SJLR outperforms SDRF \cite{topping2022understanding} and FoSR \cite{karhadkar2023fosr} in all datasets.
%Both SJLR and SDRF use the same JLC metric in Tables \ref{tbl:results_experiment_SGC} and \ref{tbl:results_experiment_GCN}, and therefore we are assessing their performance based on how the edges are added or removed.
For SJLR and SDRF, both methods use the same JLC metric in Tables \ref{tbl:results_experiment_SGC} and \ref{tbl:results_experiment_GCN}, and therefore we are assessing their performance based on how the edges are added or removed.
We argue that SJLR is a critical improvement over SDRF regarding the practical adoption of curvature-based methods in GNNs.
%Finally, we remark that some methods like GDC \cite{klicpera2019diffusion} require specific architectural changes to work properly.
Finally, we remark that some methods like GDC \cite{klicpera2019diffusion} and FA \cite{alon2021bottleneck} require specific architectural changes to work properly.
For example, we achieve the results of GDC only when applying weight decay in the first graph convolutional layer.
% as stated in Appendix \ref{app:implementation_details}.

%\begin{wrapfigure}{r}{0.33\textwidth}
%    \vspace{-10pt}
%    \includegraphics[width=0.33\textwidth]{figures/JLC_vs_BFC_SBModel_2_Clusters.eps}
%    \caption{Average running time for BFC and JLC for variations in the number of nodes.}
%    \label{fig:JLC_vs_BFC}
%    \vspace{-10pt}
%\end{wrapfigure}

\begin{figure}
    \centering
    \includegraphics[width=\columnwidth]{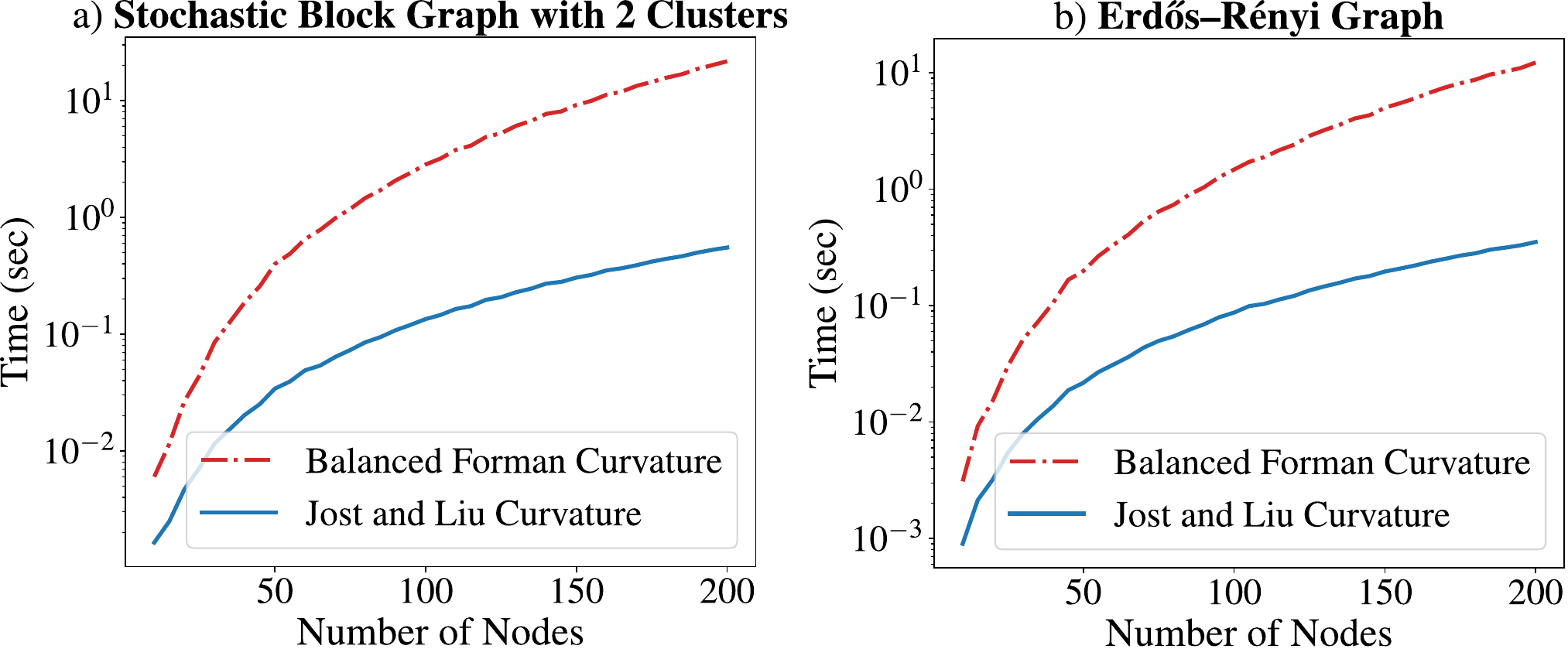}
    \caption{Average running time for BFC and JLC for variations in the number of nodes.}
    \label{fig:JLC_vs_BFC}
\end{figure}

% Table \ref{tbl:results_experiment_deep} shows the results of the experiment related to stacking multiple convolutional layers.
% We observe similar behaviors compared to the previous experiment, \ie the rewiring algorithms present state-of-the-art performance in heterophyllous datasets.
% On the other hand, methods like RDC and DGN perform better in homophilous data.
% We also notice that SJLR+ readily outperforms all previous methods in six datasets for $L=32$.
% Finally, we note that Chameleon is the only dataset where going deeper presents a benefit in performance, while in the other datasets the performance decreases.
% As a consequence, the exploration of new data besides the well-known benchmarks is a crucial practical topic for future work in over-smoothing and over-squashing \cite{dwivedi2022long,liu2022taxonomy}.
%As a consequence, the exploration of new datasets is a crucial practical topic for future work in over-smoothing and over-squashing \cite{dwivedi2022long,liu2022taxonomy}.
Figure \ref{fig:JLC_vs_BFC} shows the average running time over ten repetitions to compute the BFC and JLC metrics in a stochastic block graph and one Erdős-Rényi graph.
We notice the large gap between the computation time of the JLC and BFC, which makes JLC more suitable in practice.

%\begin{figure}
%    \centering
%    \includegraphics[width=\textwidth]{figures/JLC_vs_BFC.eps}
%    \caption{Average running time for balanced Forman curvature and Jost and Liu curvature for variations in the number of nodes in three artificial graphs.}
%    \label{fig:JLC_vs_BFC}
%\end{figure}

\subsection{Ablation Study}

We conduct an ablation study to examine the influence of the addition and removal of edges in SJLR, employing SGC as the backbone model in alignment with our theoretical findings.
To explore this, we perform hyperparameter optimization as outlined in Section \ref{sec:experiments}, and we obtain the results summarized in Table \ref{tbl:ablation}.
The findings suggest that the addition and removal of nodes are complementary and additive in performance.
For example, for nearly all datasets, incorporating the addition and removal of edges leads to better performance than performing one strategy alone.
This is an important difference regarding recent works \cite{karhadkar2023fosr,liu2023curvdrop} where only edges are added or removed.
We theoretically (Theorem \ref{trm:cheeger_constant_and_mixing_time}) and empirically (Tables \ref{tbl:results_experiment_SGC}, \ref{tbl:results_experiment_GCN}, and \ref{tbl:ablation}) show that both, removing and adding edges is required to find a good compromise in the over-smoothing over-squashing trade-off.

%In some cases, only adding edges is even harmful for the performance like in the Wisconsing dataset.
%We see that removing edges is thus fundamental for the datasets tested in this paper.

% {\color{red}
% \subsection{Ablations}

% -Doing rewiring only in the first layer, and keep same rewired graph for the whole GNN.

% -Doing only removal of edges.

% -Doing only addition of edges.
% }

\subsection{Limitations}

One of the limitations of SJLR is the optimization of hyperparameters.
Due to the expanded search space, finding an optimal set of hyperparameters for SJLR through random search becomes more challenging compared to simpler methods.
Another significant limitation is that the current implementation of SJLR relies on a bank $\mathcal{E}_a$ of good edges to add, which is based solely on the triangles of edges with the most negative curvature.
As a result, it is not possible to have edges that directly connect long-distance nodes.
A potential solution to address this limitation is to explore more sophisticated curvature metrics that are weighted with the shortest path distance between nodes, as proposed in \cite{liu2023curvdrop}.
However, implementing such an approach would introduce a high computational cost, making it impractical for large-scale graph applications.
These limitations underscore the need for further research in developing efficient curvature metrics that can effectively account for larger distances in the graph.

\section{Conclusions}
\label{sec:conclusions}

In this work, we have established a connection between the challenges of over-smoothing and over-squashing in GNNs.
We showed how both issues are intrinsically related to the spectral gap of the normalized Laplacian matrix.
Through the application of the Cheeger inequality, we have revealed the existence of a trade-off between over-smoothing and over-squashing, highlighting the inherent limitations of simultaneously addressing both problems from a topological perspective.
To tackle these challenges, we have introduced the SJLR algorithm, which utilizes a bound of the Ollivier’s Ricci curvature.
SJLR offers a computationally efficient solution compared to previous methods, such as SDRF, while still preserving essential theoretical properties.
SJLR outperformed previous methods in homophilous and heterophyllous graph datasets for node classification.
%Curvature-based rewiring algorithms, like SJLR and DE, outperformed previous methods in homophilous and heterophyllous graph datasets for node classification.
Most importantly, this work presented a crucial yet simple theoretical contribution to the fundamental problems of over-smoothing and over-squashing in GNNs.

%This work has provided significant theoretical contributions to the understanding and mitigation of over-smoothing and over-squashing in GNNs.
%However, there remain several open problems in this field that warrant further investigation.
This work opens several research directions.
%For example, exploring efficient curvature metrics, such as JLC, represents a promising direction to develop scalable and effective GNNs that can alleviate over-smoothing and over-squashing.
For example, integrating neural architecture search \cite{cai2021rethinking} into curvature-based methods could facilitate the challenge of hyperparameter optimization in the context of SJLR.\\

\noindent \textbf{Acknowledgements.} This work was supported by the DATAIA Institute as part of the ``Programme d’Investissement d’Avenir'', (ANR-17-CONV-0003) operated by CentraleSupélec, and by ANR (French National Research Agency) under the JCJC project GraphIA (ANR-20-CE23-0009-01).

\appendix

\section{Proof of Theorem \ref{trm:cheeger_constant_and_mixing_time}}
\label{app:proof_cheeger_constant_and_mixing_time}

\begin{proof}
    Let $\mathbf{P}=\mathbf{D}^{-1}\mathbf{A}$ be the random walk transition matrix, and let $f:\mathcal{V} \to \mathbb{R}$ be any initial distribution with vector representation $\mathbf{f}\in \mathbb{R}^{N\times 1}$.
    If we want to measure the distance between $\mathbf{f}^{\mathsf{T}}\mathbf{P}^s$ and the stationary distribution in the $\ell_2$ norm we need to compute $\Vert \mathbf{f}^{\mathsf{T}}\mathbf{P}^s - \boldsymbol{\pi} \Vert = \Vert \mathbf{f}^{\mathsf{T}}\mathbf{P}^s - \frac{\mathbf{1}^{\mathsf{T}} \mathbf{D}}{\vol(G)}\Vert$.
    % \begin{equation}
    %     \Vert \mathbf{f}^{\mathsf{T}}\mathbf{P}^s - \boldsymbol{\pi} \Vert = \Vert \mathbf{f}^{\mathsf{T}}\mathbf{P}^s - \frac{\mathbf{1}^{\mathsf{T}} \mathbf{D}}{\vol(G)}\Vert.
    % \end{equation}
    Let $\boldsymbol{\phi}_i$ be the orthonormal eigenfunction associated with $\lambda_i$.
    We know that $\boldsymbol{\phi}_1 = \frac{\mathbf{1}^{\mathsf{T}} \mathbf{D}^{\frac{1}{2}}}{\sqrt{\vol(G)}}$. %where $\mathbf{u}_1$ is the eigenvector corresponding to the eigenvalue $\lambda_1$. 
    Therefore, we have:
    \begin{gather}
        \nonumber
        a_1 \boldsymbol{\phi}_1 \mathbf{D}^{\frac{1}{2}} = \frac{1}{\sqrt{\vol(G)}} \frac{\mathbf{1}^{\mathsf{T}} \mathbf{D}^{\frac{1}{2}}}{\sqrt{\vol(G)}} \mathbf{D}^{\frac{1}{2}} = \frac{\mathbf{1}^{\mathsf{T}} \mathbf{D}}{\vol(G)},\\
        \text{where } a_1 = \frac{\langle \mathbf{f}^{\mathsf{T}} \mathbf{D}^{-\frac{1}{2}}, \mathbf{1}^{\mathsf{T}}\mathbf{D}^{\frac{1}{2}} \rangle}{\Vert \mathbf{1}^{\mathsf{T}} \mathbf{D}^{\frac{1}{2}} \Vert} = \frac{1}{\sqrt{\vol(G)}}.
    \end{gather}
    % \begin{equation}
    %     a_1 \boldsymbol{\phi}_1 \mathbf{D}^{\frac{1}{2}} = \frac{1}{\sqrt{\vol(G)}} \frac{\mathbf{1}^{\mathsf{T}} \mathbf{D}^{\frac{1}{2}}}{\sqrt{\vol(G)}} \mathbf{D}^{\frac{1}{2}} = \frac{\mathbf{1}^{\mathsf{T}} \mathbf{D}}{\vol(G)}, \text{ where } a_1 = \frac{\langle \mathbf{f}^{\mathsf{T}} \mathbf{D}^{-\frac{1}{2}}, \mathbf{1}^{\mathsf{T}}\mathbf{D}^{\frac{1}{2}} \rangle}{\Vert \mathbf{1}^{\mathsf{T}} \mathbf{D}^{\frac{1}{2}} \Vert} = \frac{1}{\sqrt{\vol(G)}}.
    % \end{equation}
    As a consequence we have that:
    \begin{align}
        \nonumber
        \Vert \mathbf{f}^{\mathsf{T}}\mathbf{P}^s - \frac{\mathbf{1}^{\mathsf{T}} \mathbf{D}}{\vol(G)}\Vert &= \Vert \mathbf{f}^{\mathsf{T}}\mathbf{P}^s - a_1 \boldsymbol{\phi}_1 \mathbf{D}^{\frac{1}{2}} \Vert \\
        &= \Vert \mathbf{f}^{\mathsf{T}} \mathbf{D}^{-\frac{1}{2}} (\mathbf{I}-\mathbf{L}_{\text{sym}})^s \mathbf{D}^{\frac{1}{2}} - a_1 \boldsymbol{\phi}_1 \mathbf{D}^{\frac{1}{2}}\Vert,
    \end{align}
    % \begin{equation}
    %     \Vert \mathbf{f}^{\mathsf{T}}\mathbf{P}^s - \frac{\mathbf{1}^{\mathsf{T}} \mathbf{D}}{\vol(G)}\Vert = \Vert \mathbf{f}^{\mathsf{T}}\mathbf{P}^s - a_1 \boldsymbol{\phi}_1 \mathbf{D}^{\frac{1}{2}} \Vert = \Vert \mathbf{f}^{\mathsf{T}} \mathbf{D}^{-\frac{1}{2}} (\mathbf{I}-\mathbf{L}_{\text{sym}})^s \mathbf{D}^{\frac{1}{2}} - a_1 \boldsymbol{\phi}_1 \mathbf{D}^{\frac{1}{2}}\Vert,
    % \end{equation}
    since $\mathbf{P}=\mathbf{D}^{-\frac{1}{2}} (\mathbf{I}-\mathbf{L}_{\text{sym}}) \mathbf{D}^{\frac{1}{2}}$.
    Suppose we write $\mathbf{f}^{\mathsf{T}} \mathbf{D}^{-\frac{1}{2}} = \sum_i a_i \boldsymbol{\phi}_i$, then:%where $\mathbf{u}_i$ denotes the eigenvector associated to $\lambda_i$, then:
    \begin{align}
        \nonumber
        \Vert \mathbf{f}^{\mathsf{T}}\mathbf{P}^s - \boldsymbol{\pi} \Vert &= \Vert \sum_i a_i \boldsymbol{\phi}_i (\mathbf{I}-\mathbf{L}_{\text{sym}})^s \mathbf{D}^{\frac{1}{2}} - a_1 \boldsymbol{\phi}_1 (1-\lambda_1)^s \mathbf{D}^{\frac{1}{2}} \Vert \\
        \nonumber
        &= \Vert \sum_{i \neq 1} a_i \boldsymbol{\phi}_i (1-\lambda_i)^s \mathbf{D}^{\frac{1}{2}} \Vert = \Vert \sum_{i \neq 1} (1-\lambda_i)^s a_i \boldsymbol{\phi}_i \mathbf{D}^{\frac{1}{2}} \Vert \\
        %\nonumber
        %& \leq (1-\lambda')^s \frac{\max_i \sqrt{d_i}}{\min_j \sqrt{d_j}} \\
        \nonumber
        & \leq e^{-s \lambda'} \frac{\max_i \sqrt{d_i}}{\min_j \sqrt{d_j}},\\
        \label{eqn:dist_stationary_distribution}
        &\text{where } \lambda' = \begin{cases}
            \lambda_2 & \text{if } 1-\lambda_2 \geq \lambda_N - 1, \\
            2 - \lambda_N & \text{otherwise}.
        \end{cases}
    \end{align}
    %From Section \ref{sec:stationary_distribution} we know that only $\lambda_2$ is important in \eqref{eqn:dist_stationary_distribution} since we can apply lazy walk.
    Thus, we can compute the value of $s$ such that $\Vert \mathbf{f}^{\mathsf{T}} \mathbf{P}^s - \boldsymbol{\pi} \Vert \leq \epsilon$ as follows:
    \begin{gather}
        \nonumber
        s \geq \frac{1}{\lambda' \log \left( \max_i \sqrt{d_i} / \epsilon \min_j \sqrt{d_j} \right)} \\
        \rightarrow \lambda' \geq \frac{1}{s \log \left( \max_i \sqrt{d_i} / \epsilon \min_j \sqrt{d_j} \right)}.
    \end{gather}
    % \begin{equation}
    %     s \geq \frac{1}{\lambda' \log \left( \max_i \sqrt{d_i} / \epsilon \min_j \sqrt{d_j} \right)} \rightarrow \lambda' \geq \frac{1}{s \log \left( \max_i \sqrt{d_i} / \epsilon \min_j \sqrt{d_j} \right)}.
    % \end{equation}
    In general, suppose that we have a weighted graph with weights $w(u,v)$, and so we have eigenvalues $\lambda_i$ with $\lambda_N - 1 \geq 1 - \lambda_2$.
    Therefore, we can modify the weights with some constant $c$ as follows:
    \begin{equation}
        w'(u,v) = \begin{cases}
            w(v,v) + cd_v & \text{if } u=v, \\
            w(u,v) & \text{otherwise}.
        \end{cases}
    \end{equation}
    Notice that the lazy walk is given by $c=1$.
    The weighted graph has eigenvalues:
    \begin{equation}
        \lambda_k’ = \frac{\lambda_k}{1+c} = \frac{2\lambda_k}{\lambda_N + \lambda_2}, \text{ where } c=\frac{\lambda_2 + \lambda_N}{2} -1 \leq \frac{1}{2}.
    \end{equation}
    We thus have $1-\lambda_2’ = \lambda_N’ - 1 = \frac{\lambda_N-\lambda_2}{\lambda_N+\lambda_2}$.
    Since $c \leq \frac{1}{2}$ and we have that $\lambda_k’ \geq \frac{2\lambda_k}{2+\lambda_k} \geq \frac{2}{3}\lambda_k$ for $\lambda_k \leq 1$.
    Particularly, we can set:
    \begin{equation}
        \lambda=\lambda_2’ = \frac{2\lambda_2}{\lambda_N+\lambda_2} \geq \frac{2}{3}\lambda_2.
    \end{equation}
    Therefore, the modified random walk corresponding to the weighted function $w’$ has an improved bound for the convergence rate in $\ell_2$ distance as follows:
    \begin{equation}
        \frac{1}{\lambda} \log \left( \frac{\max_i \sqrt{d_i}}{\epsilon \min_j \sqrt{d_j}} \right).
    \end{equation}
    Please see \cite{chung1997spectral} for further details.
    Finally, using the Cheeger inequality in \eqref{eqn:cheeger_inequality} we have that:
    \begin{equation}
        2h_G \geq \lambda_2 \geq \frac{1}{s} \log \left( \frac{\max_i \sqrt{d_i}}{\epsilon \min_j \sqrt{d_j}} \right) \rightarrow 2h_G \geq \frac{1}{s} \log \left( \frac{\max_i \sqrt{d_i}}{\epsilon \min_j \sqrt{d_j}} \right).
    \end{equation}
\end{proof}

\section{Mixing Time and Cheeger Constant}
\label{app:mixing_time}

The concepts of random walks, Cheeger constant, and convergence to the stationary distribution are tightly related to Markov chains, conductance (Cheeger constant), and the mixing time.
We can represent any Markov chain as a random walk on some weighted directed graph $G$.
%Let $\mathbf{P}$ be the transition matrix for the directed graph associated with an ergodic Markov chain.
The mixing time $\tau(\epsilon)$ of an ergodic Markov chain is the time until the Markov chain is close to its stationary distribution, \ie $s$. We can find an upper bound for $\tau(\epsilon)$ as follows:
\begin{theorem}[Sinclair \cite{sinclair2012algorithms}]
\label{trm:mixing_time}
Let $h_G$ be the Cheeger constant of an ergodic and reversible Markov chain.
For $\epsilon > 0$ we have that:
\begin{equation}
    s = \tau(\epsilon) \leq \frac{2}{h_G^2}\left( \log\left( \frac{1}{\epsilon} \right) + \log\left( \frac{1}{\pi_*} \right)  \right),
\end{equation}
Where $\pi_*=\min_{j \in \Omega} \boldsymbol{\pi}(j)$, and $\Omega$ is the state space of the Markov chain.
%Proof: See \cite{sinclair2012algorithms}.
\end{theorem}
From Theorem \ref{trm:mixing_time}, and since $h_G > 0$ and $\lambda_2 > 0$ for connected graphs, we have that $\lim_{h_G \to \infty} \frac{2}{h_G^2}\left( \log\left( \frac{1}{\epsilon} \right) + \log\left( \frac{1}{\pi_*} \right)  \right) = 0$ so that if $h_G \to \infty$ then $s \to 0$.
Similarly, we have that $\lim_{s \to \infty} \frac{2}{s}\left( \log\left( \frac{1}{\epsilon} \right) + \log\left( \frac{1}{\pi_*} \right)  \right) = 0$, so that if $s \to \infty$ then $h_G^2 \to 0$.
Finally, considering the results from Theorem \ref{trm:cheeger_constant_and_mixing_time} we have that $s \to \infty \iff h_G \to 0$ and $h_G \to \infty \iff s \to 0$.

\section{Sweet Point}
\label{app:sween_point}

We can propose an optimization problem to find the ``sweet point'' in the over-smoothing vs. over-squashing trade-off as follows:
\begin{equation}
    \argmin_{\mathcal{E}' \in \mathcal{V} \times \mathcal{V}} h_{G'} + \rho f(\lambda_2',\epsilon),
    \label{eqn:combinatorial_optimization_problem}
\end{equation}
where $G'=(\mathcal{V},\mathcal{E}')$, $\rho$ is a regularization parameter, $f(\lambda_2',\epsilon)$ is the function of mixing steps of $G'$, $\lambda_2'$ is the spectral gap of $G'$, and $\epsilon$ is a hyperparameter.
The optimization problem in \eqref{eqn:combinatorial_optimization_problem} poses several challenges:
\begin{enumerate}[leftmargin=*]
    \item Since $\mathcal{E}' \in \mathcal{V} \times \mathcal{V}$, the optimal solution could be very different from $\mathcal{E}$, destroying the original structural information of the graph.
    \item Computing the Cheeger constant of a finite graph is an NP-hard problem \cite{garey1974some}, and the algorithm to solve \eqref{eqn:combinatorial_optimization_problem} could require calculating $h_{G'}$ several times.
    \item The optimization of the hyperparameters $\rho$ and $\epsilon$ should be data-driven for the specific graph learning task. 
\end{enumerate}
As a consequence, solving the combinatorial optimization problem in \eqref{eqn:combinatorial_optimization_problem} is practically infeasible.
Instead of trying to solve \eqref{eqn:combinatorial_optimization_problem} directly, we have proposed SJLR as a heuristic algorithm.
In SJLR, we relate to the Cheeger constant indirectly through Corollaries \ref{cor:relationship_ricci_cheeger_constant} and \ref{cor:JLC_positive} with the JLC metric.
Similarly, we relate to the mixing step $f(\lambda_2',\epsilon)$ through the Lemma \ref{lem:upper_bound_lambda_2} when dropping edges in Algorithm \ref{alg:SJLR}.

\bibliographystyle{ACM-Reference-Format}
\bibliography{bibfile}

\end{document}